\pdfoutput=1
 \documentclass[12pt, a4paper, oneside]{article}

\usepackage{latexsym}
\usepackage{amssymb}
\usepackage{mathrsfs}
\usepackage{graphicx}
\usepackage[latin1]{inputenc}
\usepackage{natbib}
\usepackage{multirow}
\usepackage{url}
\usepackage{dcolumn}
\usepackage{amsthm}
\usepackage{amsmath}
\usepackage{array}
\usepackage{amsfonts}
\usepackage{bm}
\usepackage[german]{varioref}
\usepackage{psfrag}
\usepackage{caption}
\usepackage{rotating}
\usepackage{oldgerm}
\usepackage{enumerate}
\usepackage{hyperref}
\usepackage{here}	
\usepackage{pdflscape}
\usepackage{listings}
\usepackage{subcaption}
\usepackage[english]{babel}
\usepackage[normalem]{ulem}
\usepackage[lined,boxed,linesnumbered]{algorithm2e}
\usepackage{color}
\usepackage{footmisc}
\usepackage{enumitem}
\usepackage{dsfont} 
\usepackage{tabularx}

\RequirePackage[normalem]{ulem} 
\RequirePackage{color}\definecolor{RED}{rgb}{1,0,0}\definecolor{BLUE}{rgb}{0,0,1} 

\newcolumntype{L}[1]{>{\raggedright\arraybackslash}p{#1}} 
\newcolumntype{C}[1]{>{\centering\arraybackslash}p{#1}} 
\newcolumntype{R}[1]{>{\raggedleft\arraybackslash}p{#1}} 

\long\def\symbolfootnote[#1]#2{\begingroup%
\def\thefootnote{\fnsymbol{footnote}}\footnote[#1]{#2}\endgroup}

\numberwithin{equation}{section}

\newtheorem{Theorem}{Theorem}[section]

\newtheorem{example}[Theorem]{Example}

\allowdisplaybreaks[1]
\newcommand{\CCC}{\mathcal{C}}

\newcommand{\HHH}{\mathcal{H}}
\newcommand{\FFF}{\mathcal{F}}
\newcommand{\GGG}{\mathcal{G}}
\newcommand{\III}{\mathcal{I}}

\newcommand{\LLL}{\mathcal{L}}
\newcommand{\NNN}{\mathcal{N}}

\newcommand{\UUU}{\mathcal{U}}
\newcommand{\VVV}{\mathcal{V}}

\newcommand{\coloneqq}{:=}

\DeclareMathOperator*{\argmax}{arg\,max}
\DeclareMathOperator{\tr}{tr}																
\newcommand{\separate}[4]
           {#1 \mathrel{\perp} #2 \mathrel{|} #3 \; \left[ #4 \right]}
\newcommand{\indep}{\mathrel{{\perp}\hspace*{-0.6em}{\perp}}}
\newcommand{\given}{\mathrel{|}}
\newcommand{\condindep}[3]{#1 \indep #2 \given #3}

\DeclareMathOperator{\AIC}{AIC}

\begin{document}
\title{\textbf{Dependence Modeling in Ultra High Dimensions with Vine Copulas and the Graphical Lasso}}
\author{Dominik M\"uller\thanks{Corresponding author} \thanks{Department of Mathematics, Technische Universit\"at M\"unchen,
		Boltzmannstraße 3, 85748 Garching, Germany. E-Mail: \href{mailto:dominik.mueller@ma.tum.de}{dominik.mueller@ma.tum.de}, \href{mailto:cczado@ma.tum.de}{cczado@ma.tum.de}.}
	\quad
	Claudia Czado\footnotemark[2]}
  \maketitle
  
\begin{abstract}
	To model high dimensional data, Gaussian methods are widely used since they remain tractable and yield parsimonious models by imposing strong assumptions on the data. Vine copulas are more flexible by combining arbitrary marginal distributions and (conditional) bivariate copulas. Yet, this adaptability is accompanied by sharply increasing computational effort as the dimension increases. The approach proposed in this paper overcomes this burden and makes the first step into ultra high dimensional non-Gaussian dependence modeling by using a divide-and-conquer approach. First, we apply Gaussian methods to split datasets into feasibly small subsets and second, apply parsimonious and flexible vine copulas thereon. Finally, we reconcile them into one joint model. We provide numerical results demonstrating the feasibility of our approach in moderate dimensions and showcase its ability to estimate ultra high dimensional non-Gaussian dependence models in thousands of dimensions.
\end{abstract}


\section{Introduction}\label{sec:introduction}
In many areas of scientific research but also in business applications, high dimensional problems arise. For example, if a financial institution owns stocks $S_1,\dots,S_{100}$ and wants to calculate a \textit{portfolio Value-at-Risk}, see e.\,g.\ \citet{McNeilBook}, a joint dependence model in $d=100$ dimensions is required. For large companies, one can easily imagine this number to increase into thousand for a single asset class and much higher when several asset classes are considered. Another active field of research in which dimensions grow rapidly is computational biology. For example in \textit{gene expression data}, genes are measured simultaneously to make inference about dependence within the biological system or with respect to some disease, see e.\,g.\ \citet{toh2002inference}. In this case, the number of genes measured describes the dimension $d$ and in recent applications can increase to several thousands. Similar to this, hundreds of \textit{metabolites} in the human blood can be measured and analyzed for dependence, see e.\,g.\ \citet{krumsiek2011gaussian}, where $d = 151$. This example is also easily imaginable to be extended to thousands of different metabolites.\\
Among the most prominently used models to analyse such datasets are \textit{Gaussian Graphical Models}, which are based on the assumption that the data originates from a multivariate Gaussian distribution in $d$ dimensions. Neglecting the mean of the distribution, the problem remains to estimate a $d \times d$ covariance matrix $\Sigma$.
A favourable property of the multivariate Gaussian distribution is that for $\Omega = \Sigma^{-1}$ and a zero entry of $\Omega$, i.\,e.\ $\omega_{ij} = 0$, we have conditional independence of $i$ and $j$ given the rest $\left\{1,\dots,d\right\} \setminus \left\{i,j\right\}$. By drawing a graph with nodes $\left\{1,\dots,d\right\}$ and omitting an undirected edge $\left(i,j\right)$ whenever $\omega_{ij} = 0$, we obtain a \textit{graphical model} for the conditional independence in this distribution. There has been a considerable effort on how to estimate (sparse) inverse covariance matrices and thus, Gaussian graphical models. It started with \textit{covariance selection} of \citet{Dempster1972} to the current state of the art algorithm, the \textit{graphical Lasso} \citep{glasso}. The huge advantage of these methods is their computational tractability also in ultra high dimensions, i.\,e.\ several thousands of variables. The underlying assumption of Gaussianity is however very strict, by imposing also assumptions on the marginals. This has been relaxed by the so called \textit{non-paranormal} \citet{liu2012} where the marginal distributions must not necessarily be Gaussian. For the dependence part however, especially for financial data, the multivariate Gaussian is a too strong simplification. It does not allow for heavy tails as the Student's-t distribution or asymmetric dependence, thus more sophisticated models are required.\\
The main idea of the so called \textit{pair copula construction (PCC)}, see \citet{Aasetal2009}, is to couple $d(d-1)/2$ (conditional) bivariate distributions and $d$ marginal distributions to obtain a joint $d$-dimensional distribution, called \textit{vine copula}. The huge benefit is now that all involved distributions can be chosen arbitrarily and entirely independently from each other. However, this construction is not unique but is described by a graphical model, a \textit{regular (or R-)vine} \citep{BedfordCooke2001, BedfordCooke2002}. Each possible R-vine constitutes a different construction of a $d$-dimensional distribution. As the number of possible R-vines grows super-exponentially in dimensions, a search for an optimal model in terms of a goodness of fit criteria such as \textit{log-likelihood} is not feasible in any dimension, and heuristic algorithms as the one of \citet{dissmann-etal} come into place. Secondly, an R-vine model needs $d(d-1)/2$ pair copulas. Clearly, this is growing too fast in hundreds of dimensions, unspoken of thousands. Thus, this also demands sophisticated approaches in high dimensions to keep models parsimonious. The large number of applications with vine copulas in recent years shown in \citet{econometrics4040043} can be easily thought to expand to hundreds and thousands of dimensions, which makes an extension of vine copulas in ultra high dimensions desirable.\\
Our contribution is twofold. First, we will show how well established Gaussian methods in high dimensions can be fundamental for clustering R-vines. Thus, we break a $d$-dimensional dependence model in multiple dependence sub-models with significantly smaller dimensions. These sub-models are now tractable again. Secondly, within these sub-models, we use a refined algorithm to improve the accuracy of the standard search algorithm for vine copulas by \citet{dissmann-etal}. Afterwards, the sub-models are recombined to obtain one joint parsimonious model in ultra high dimensions. We show that this is working well in moderate dimensions and outperforming previous methods in several hundreds of dimensions in computation time and goodness of fit. Going to ultra high dimensions, i.\,e.\ several thousands, it is to our knowledge the only feasible way to estimate vine copula models, and is actually doable in a comparably short amount of time. Finally, we will also demonstrate that non-Gaussian models give clearly a competitive edge on Gaussian models using real world financial data.\\
The paper is structured as follows. We will briefly introduce vine copulas in Section \ref{sec:rvines} and discuss current model selection methods. We recapitulate graphical models based on the multivariate Gaussian distribution in Section \ref{sec:undirectedgraphicalmodels} where we focus on the graphical Lasso. Section \ref{sec:selectvines} contains our \textit{divide-and-conquer} approach where we split the R-vine selection into sub-problems according to a path of solutions of the graphical Lasso and solve each separately with increased accuracy. We sketch an algorithmic implementation and continue to numerical examples in Section \ref{sec:numericalexamples}. There, we show a simulation study in $85$ dimensions to demonstrate the feasibility of our approach in moderate dimensions. Afterwards, we increase the dimension to over $1700$ to demonstrate the high efficiency of our approach with respect to time consumption, at the same time outperforming standard methods in terms of penalized goodness of fit measures. We finally include an example in more than $2000$ dimensions which demonstrates that a pure Gaussian fitting is too restrictive for real world datasets, and can be improved significantly by ultra high dimensional vine copulas.

\section{Dependence Modeling with Vine Copulas}\label{sec:rvines}
We will use the following conventions. Upper case letters $X$ denote random variables, and lower case letters $x$ their realizations. We use bold letters $\bm{X}$ for random vectors and $\bm{x}$ for the vector of realizations. Matrices $M$ are identified by upper case letters. We write $v_i$ for the $i$-th entry of the vector $\bm{v}$. Denote sub vectors of $\bm{x}=\left(x_1,\ldots,x_d\right)^T$ by $\bm{x}_{D\left(e\right)} \coloneqq \left(\bm{x}_j\right)_{j \in D\left(e\right)}$. When considering matrices, we denote $m_{i,j}$ the $j$-th entry in the $i$-th row of the matrix $M$. Additionally, we use the following data scales.
\begin{enumerate}\label{en:scales}
	\itemsep-0.25em
	\item \textit{x-scale}: the original scale of $X_i$, i.i.d., with density $f_i(x_i),\ i=1,\dots,d$,
	\item \textit{u-scale} or \textit{copula-scale}: $U_i = F_i\left(X_i\right)$, $F_i$ the cdf of $X_i$ and $U_i \sim \UUU\left[0,1\right]$, $i=1,\dots,d$,
	\item \textit{z-scale}: $Z_i = \Phi^{-1}\left(U_i\right)$, $\Phi$ the cdf of $\NNN\left(0,1\right)$ thus $Z_i \sim \NNN\left(0,1\right)$, $i=1,\dots,d$.
\end{enumerate}
For a random vector $\bm{X} = \left(X_1,\ldots,X_d\right)$ we denote the joint distribution function and density by $F$ and $f\left(x_1,\dots,x_d\right) = \frac{\partial F}{\partial x_1,\dots,\partial x_d}\left(x_1,\dots,x_d\right)$, respectively. To model $F$, we exploit the famous Theorem of \citet{Sklar1959}. It separates the marginal distributions from the joint distribution such that $F\left(x_1,\ldots,x_d\right) = C\left(F_1\left(x_1\right),\ldots,F_d\left(x_d\right)\right)$, where $C$ denotes a $d$-dimensional \textit{copula}. If all marginal distributions $F_i$ are continuous, $C$ is unique. The corresponding density $f$ with respect to the copula is obtained by taking derivatives
\begin{equation}\label{eq:fdensity}
f\left(x_1,\ldots,x_d\right) = c\left(F_1\left(x_1\right),\ldots,F_d\left(x_d\right)\right) \prod_{i=1}^d~f_i\left(x_i\right),
\end{equation}
with $c\left(x_1,\dots,x_d\right)=\frac{\partial C}{\partial x_1,\dots, \partial x_d}\left(x_1,\dots,x_d\right)$ the \textit{copula density}. While there exist many model classes for the univariate marginal distributions, this is not the case for copulas in arbitrary dimensions. The most well known copula functions are the multivariate Gaussian, Student $t$, Archimedean and extreme value copulas. Using these copulas to model $d$-dimensional data comes along with several drawbacks such as lack of flexibility and computational issues. To overcome this, \citet{Joe1996} constructed distributions in $d$ dimensions modelled by $d(d-1)/2$ bivariate distributions. Extending his work, \citet{Aasetal2009} developed the \textit{pair-copula-construction (PCC)}, which builds up a $d$-dimensional distribution using $d$ marginals and $d(d-1)/2$ (conditional) bivariate copulas. These building blocks can be chosen entirely independent from each other and thus provide a very flexible modeling approach. For example, pair copulas with heavy tails or asymmetric dependence can be used. Yet, the construction of a $d$-dimensional distribution with $d(d-1)/2$ (conditional) pairs is not unique. More precisely, there exist an exponentially growing number of valid constructions, see \citet[p.\ 190]{kuro:joe:2010}. With the work of \citet{BedfordCooke2001, BedfordCooke2002}, introducing \textit{regular vines}, a framework was developed which allowed to organize the possible constructions by \textit{vine trees}. In total, $d-1$ of these trees are required to define a $d$-dimensional distribution and are given by $\VVV = \left(T_1,\ldots,T_{d-1}\right)$ such that
\begin{enumerate}[label={(\roman*})]
	\itemsep0em 
	\item $T_1$ is a tree with nodes $V_1=\left\{1,\ldots,d\right\}$ and edges $E_1$,
	\item for $i \ge 2$, $T_i$ is a tree with nodes $V_i = E_{i-1}$ and edges $E_i$,
	\item if two nodes in $T_{i+1}$ are joined by an edge, the corresponding edges in $T_i$ must share a common node (proximity condition).
	\label{eq:proximitycondition}
\end{enumerate}
By (ii), edges become nodes and are connected with new edges recursively. For a node, e.\,g.\ $\left\{1,2\right\} \in V_2$, we define the two nodes $\left\{1\right\}, \left\{2\right\} \in V_1$ of which the node in $V_2$ is combined, as \textit{m-children}. For some node in $T_k$, define the \textit{m-family} as the union of all its m-children and their m-children in trees $T_1,\dots,T_{k-1}$. Each edge in one of the R-vine trees consists of a bivariate \textit{conditioned set} and a \textit{conditioning set}, ranging from the empty set to a set containing $d-2$ variables. To specify how an edge represents a specific (conditional) pair, let the \textit{complete union} of an edge $e$ be $A_e \coloneqq \left\{j \in V_1|\exists \ e_1 \in E_1,\ldots,e_{i-1}\in E_{i-1}: j \in e_1 \in \ldots \in e_{i-1} \in e\right\}$. The conditioning set of an edge $e=\left\{a,b\right\}$ is then given by $D_e \coloneqq A_a \cap A_b$. The conditioned set is given by $\CCC_e \coloneqq \CCC_{e,a} \cup \CCC_{e,b} \mbox{ with } \CCC_{e,a} \coloneqq A_a \setminus D_e \mbox{ and } C_{e,b} \coloneqq A_b \setminus D_e$. For all edges $e \in E_i,\ 1 \le i \le d-1$, we define the set of bivariate copula densities by $\mathcal{B}\left(V\right) = \left\{c_{j\left(e\right),\ell\left(e\right);D\left(e\right)}|e \in E_i, 1 \le i \le d-1\right\}$ with the conditioned set $j\left(e\right),\ell\left(e\right)$ and the conditioning set $D\left(e\right)$. Hence, with the PCC, Equation \eqref{eq:fdensity} can be written as
\begin{equation}\label{eq:vinedensity}
f\left(x_1,\ldots,x_d\right) = \left(\prod_{i=1}^{d}~f_i\left(x_i\right)\right)
\times \left(\prod_{i=1}^{d-1}~\prod_{e\in E_i}~c_{j\left(e\right),\ell\left(e\right);D\left(e\right)}\bigg(F\left(x_{j\left(e\right)}|\bm{x}_{D\left(e\right)}\right),F\left(x_{\ell\left(e\right)}|\bm{x}_{D\left(e\right)}\right)\bigg)\right).
\end{equation}
In \eqref{eq:vinedensity}, we implicitly took into account the \textit{simplifying assumption}. It imposes that a two-dimensional conditional copula density, e.\,g.\ 
\begin{equation*}
c_{13;2}\left(F_{1|2}\left(x_1|x_2\right),F_{3|2}\left(x_3|x_2\right);x_2\right)
\end{equation*}
is independent of the conditioning value $X_2 = x_2$. A detailed discussion can be found in \citet{stoeber-vines}.
We define the parameters of the bivariate copula densities $\mathcal{B}\left(V\right)$ by $\theta\left(\mathcal{B}\left(V\right)\right)$. This determines the R-vine copula $\left(V,\mathcal{B}\left(V\right),\theta\left(\mathcal{B}\left(V\right)\right)\right)$. An intuitive representation of vine copulas is given by lower triangular $d\times d$ matrices, see \citet{dissmann-etal}. Such an R-vine matrix $M = \left(m_{i,j}\right)_{i=1,\dots,d; j=1,\dots,d}$ has to satisfy three properties.
\begin{enumerate}[label={(\roman*})]
	\itemsep0em 
	\item $\left\{m_{d,i},\dots,m_{i,i}\right\} \subset \left\{m_{d,j},\dots,m_{j,j}\right\}$ for $1 \geq i \ge j \geq d$,
	\item $m_{i,i} \notin \left\{m_{i+1,i+1},\dots,m_{d,i+1}\right\}$,
	\item for all $j=d-2,\dots,1$, $i=j+1,\dots,d$, there exist $\left(k,\ell\right)$ with $k < j$ and $\ell < k$ such that
	\begin{equation}\label{eq:proximitycondition3}
	\begin{aligned}
	\left\{m_{i,j}, \left\{m_{d,j},\dots,m_{i+1,j}\right\}\right\} &= \left\{m_{k,k}, \left\{m_{1,k},\dots,m_{\ell,k}\right\}\right\} \mbox{ or } \\
	\left\{m_{i,j}, \left\{m_{d,j},\dots,m_{i+1,j}\right\}\right\} &= \left\{m_{\ell,k}, \left\{m_{1,k},\dots,m_{\ell-1,k},m_{k,k}\right\}\right\}.
	\end{aligned}
	\end{equation}
\end{enumerate}
The last property reflects the proximity condition. We now give an example R-vine.
\begin{example}[R-vine in 6 dimensions]\label{ex:exvine1}
	The R-vine tree sequence in Figure \ref{fig:exvine1:1} is described by the R-vine matrix $M$ as follows. Edges in $T_1$ are pairs of the main diagonal and the lowest row, e.\,g.\ $\left(2{,}1\right)$, $\left(6{,}2\right)$, $\left(3{,}6\right)$. $T_2$ is given by the main diagonal and the second last row conditioned on the last row, e.\,g.\ $6{,}1|2$; $3{,}2|6$. Higher order trees are characterized similarly. M-children of e.\,g.\ $6{,}1|2$ are $\left(2{,}1\right)$ and $\left(6{,}2\right)$ and its m-family comprises additionally $1,2,6$. For a column $p$ in $M$, only entries of the main-diagonal right of $p$, i.\,e.\ values in $\left(m_{p+1,p+1},\dots,m_{d,d}\right)$ are allowed and no entry occurs more than once in a column.\\
	\begin{tabular}{cc}
	\begin{minipage}[l]{0.25\textwidth}
		\begin{equation*}
		\left(
		\begin{array}{cccccc}
		4&&&&&\\
		1&5&&&&\\
		3&1&3&&&\\
		6&3&1&6&&\\
		2&6&2&1&2&\\
		5&2&6&2&1&1
		\end{array}
		\right)
		\end{equation*}
		\begin{center}
			R-vine matrix $M$.
		\end{center}
	\end{minipage}
	&
	\begin{minipage}[r]{0.75\textwidth}
		\centering
		\includegraphics[width=0.35\textwidth]{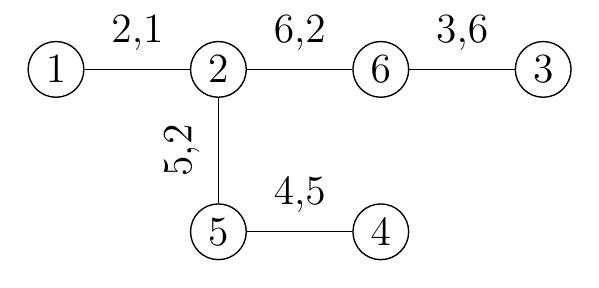}
		\includegraphics[width=0.35\textwidth]{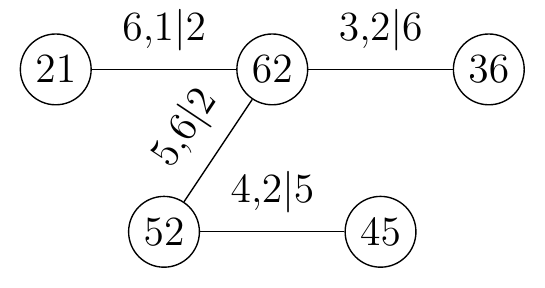}\\
		\includegraphics[width=0.35\textwidth]{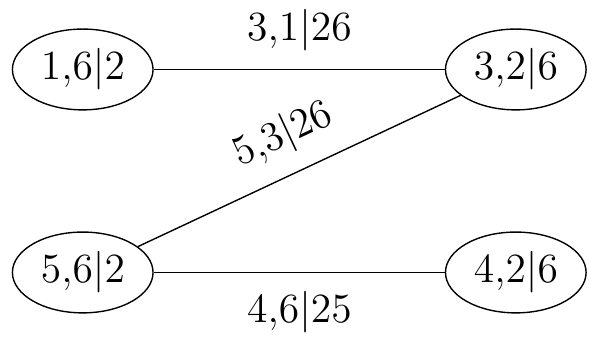}
		\includegraphics[width=0.25\textwidth]{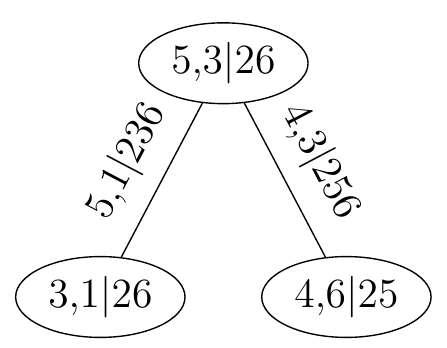}
		\includegraphics[width=0.3\textwidth]{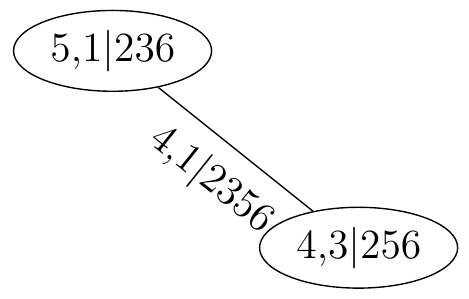}
		\captionof{figure}{R-vine trees $T_1, T_2$ (top), $T_3, T_4, T_5$ (bottom), left to right.}
		\label{fig:exvine1:1}
	\end{minipage}
\end{tabular}
	We abbreviate $c_{j,\ell|D} \coloneqq c_{j,\ell;D}\left(F\left(x_i|\bm{x}_	D\right),F\left(x_j|\bm{x}_D\right)\right)$ for the conditioning vector $\bm{x}_D$. Additionally, let $\bm{x} = \left(x_1,\dots,x_{6}\right)$ and $f_i \coloneqq f_i(x_i)$. The density in \eqref{eq:vinedensity} for this case becomes
	\begin{equation*}
	\begin{aligned}
	f\left(\bm{x}\right) = & f_1 \times f_2 \times f_3 \times f_4 \times f_5 \times f_6 \times c_{2,1} \times c_{6,2} \times c_{3,6} \times c_{5,2}\times c_{4,5} \times c_{6,1|2} \times c_{3,2|6} \\
	&\times c_{5,6|2} \times c_{4,2|5} \times c_{3,1|26} \times c_{5,3|26} \times c_{4,6|25} \times c_{5,1|236} \times c_{4,3|256}  \times c_{4,1|2356}.
	\end{aligned}
	\end{equation*}
	The pair copula families and parameters are also described by lower triangular family and parameter matrices $\Gamma = \left(\gamma_{i,j}\right)_{i=1,\dots,d; j=1,\dots,d}$ and $P = \left(p_{i,j}\right)_{i=1,\dots,d; j=1,\dots,d}$. Thus, the family and parameters of the edge $6,1|2$ are given by $\gamma_{5,4}$ and $p_{5,4}$, respectively. When two-parametric pair copulas are considered, we use an additional parameter matrix $P_2$.
\end{example}

\subsection{Model Assessment}
We consider an R-vine model in $d$ dimensions with specification $\bm{\Theta} =  \left(V,\mathcal{B}\left(V\right),\theta\left(\mathcal{B}\left(V\right)\right)\right)$. Additionally, assume we have $n$ replications of $d$-dimensional data $\left(\bm{x}_1,\dots,\bm{x}_n\right)^T \in \mathbb{R}^{n \times d}$ with $\bm{x}_k \in \mathbb{R}^d$ for $k=1,\dots,n$. Including the marginal distributions $f_i\left(x_i\right),\ i=1,\dots,d$, the log-likelihood for the specification $\Theta$ on the \textit{u-scale} is 
\begin{equation*}
\begin{aligned}
\LLL\left(\bm{\Theta},\left(\bm{x}_1,\dots,\bm{x}_n\right)\right) = 
& \sum_{k=1}^n~\Biggl(\sum_{i = 1}^d~\log\Big(f_i\left(x_{k,i}\right)\Big) + \\
& \sum_{i=1}^{d-1}~\sum_{e\in E_i}~ \log\bigg(c_{j\left(e\right),\ell\left(e\right);D\left(e\right)}\Big( F\left(x_{k,j\left(e\right)}|\bm{x}_{k,D\left(e\right)}\right), F\left(x_{k,\ell\left(e\right)}|\bm{x}_{k,D\left(e\right)}\right)\Big)\biggr)\Biggr)
\end{aligned}
\end{equation*}
The log-likelihood will always increase whenever more parameters are included in a model. Thus, in high dimensional setups where the number of significant parameters grows slower than the total number of possible parameters, it is not feasible to use log-likelihood for model discrimination. Because of this, \textit{penalized goodness of fit} measures as the (AIC) and the \textit{Bayesian information criterion (BIC)} \citep{Schwarz1978} were developed. For $n \geq 8$, \textit{BIC} will penalize more than \textit{AIC}. If the number of possible parameters in an R-vine $q\left(d\right) = 2 \times d\left(d-1\right)/2$ is greater or equal than the sample size and the model is comparably small, BIC is no longer consistent and will penalize too little. For these setups, we use the \textit{generalized information criterion (GIC)}, see \citet{FanTangTuning2013}. More precisely, we have
\begin{equation}\label{eq:AICBICGIC}
\begin{aligned}
AIC\left(\bm{\Theta},\left(\bm{x}_1,\dots,\bm{x}_n\right)\right) &= -2 \LLL\left(\bm{\Theta},\left(\bm{x}_1,\dots,\bm{x}_n\right)\right) + 2p\\
BIC\left(\bm{\Theta},\left(\bm{x}_1,\dots,\bm{x}_n\right)\right) &= -2 \LLL\left(\bm{\Theta},\left(\bm{x}_1,\dots,\bm{x}_n\right)\right) + \log\left(n\right)p,\\
GIC\left(\bm{\Theta},\left(\bm{x}_1,\dots,\bm{x}_n\right)\right) &= -2 \LLL\left(\bm{\Theta},\left(\bm{x}_1,\dots,\bm{x}_n\right)\right) + \log\left(\log\left(n\right)\right)\log\left(p\right)p,
\end{aligned}
\end{equation}
where $p$ equals the number of parameters in the model $\bm{\Theta}$. Finally note that the number of parameters correspond to the the number of one parametric copulas and two times the number of two parametric copulas in the model.

\subsection{Model Selection}\label{subsec:modelselection}
Since the space of all R-vine structures is too large to explore explicitly each model, the standard estimation method relies on using a search heuristic such as the \textit{Di{\ss}mann algorithm} \citep{dissmann-etal}. Initially, for each pair $\left(j,\ell\right) \in \binom{d}{2}$, Kendall's $\tau$ of the pair $\left(U_j,U_\ell\right)$ is calculated. The intuition is that variable pairs with high dependence should contribute significantly to the model fit and should be included in the first trees. Since the $T_1$ must be a tree, the $d-1$ edges with highest sum of absolute value of Kendall's $\tau$ are chosen based on a maximum spanning tree algorithm, e.\,g.\ \citet{Prim1957}. Afterwards, on the selected edges either maximum likelihood estimation for all desired pair copula types is performed or the corresponding copula parameters are estimated by inversion of the empirical Kendall's $\tau$. The later is only possible for one-parametric pair copula families. From these estimators, \textit{pseudo-observations} are generated. More precisely, assume we want to estimate the pair copula density $c_{j\left(e\right),\ell\left(e\right);D\left(e\right)}$. Then, we use the pseudo-observations generated from $\widehat{F}_{j|D}\left(x_{j\left(e\right)}|\bm{x}_{D\left(e\right)}\right)$ and $\widehat{F}_{\ell|D}\left(x_{\ell\left(e\right)}|\bm{x}_{D\left(e\right)}\right)$ to estimate pair copula families and parameters on this pair. The corresponding $\widehat{F}_{j|D}$ and $\widehat{F}_{\ell|D}$ are given explicitly by derivatives of the pair copula distribution functions  in lower trees evaluated at the estimated parameters. After taking into account the proximity conditions, Kendall's $\tau$ is calculated on all admissible pairs of pseudo-observations and again, a maximum spanning tree is calculated. After $d-1$ iterations, the R-vine structure is determined. This proceeding has some drawbacks. First, it is not ensured that for each tree the maximum spanning tree in terms of Kendall's $\tau$ leads to a structure with e.\,g.\ optimal log-likelihood. Second, as lower order trees influence higher order trees, sub-optimal choices lead to error-propagation. Finally, in each step a tree is fitted over all remaining (conditional) pairs, and clusters within dependence are not treated any different from structures with less dependence. Overall the effort is of order $d^2$ since $d\left(d-1\right)/2$ pair copulas are estimated.\\
As mentioned, using Kendall's $\tau$ as edge weights is only heuristically backed by the goal to model the strongest dependency first. However, this approach is not ensured to optimize e.\,g.\ log-likelihood or AIC in a given tree and thus, also not in the entire R-vine. We can instead also estimate pair copula densities $c_{j,\ell}$ for each edge $\left(j,\ell\right)$ on all $d(d-1)/2$ edges in the first tree and then calculate a maximum spanning tree with respect to an edge weight $\mu_{j,\ell}$, e.\,g.\ log-likelihood or AIC, based on the actual fit. Thus, this choice would optimize for e.\,g.\ log-likelihood or AIC in a given tree. For higher trees, this can be done similarly. This approach has firstly been discussed in \citet{rvineselect}. However, it was considered having unacceptable computational complexity. We will however come back to it later for an improved version.\\
The algorithm proposed by \citet{MuellerCzado2016} uses graphical models, more precisely \textit{directed acyclic graphs (DAGs)}, to find parsimonious structures and set the majority of pair copulas to the independence copula in larger datasets, which eases the computational effort. However, for both their and Di{\ss}mann's algorithm, more than $500-1000$ dimensions are not solvable because of the quadratically increasing effort in terms of computation time and memory.\\
Another promising proposal has been made by \citet{KrausCzado2017},  explicitly searching for simplified R-vine structures. One can expect that the true structure of the R-vine can be correctly identified if the data is originating from a simplified R-vine. However, it works similar to Di{\ss}mann's algorithm and hence, has the same computational complexity.\\
A different approach was proposed in \citet{MuellerCzado2017}, where the Lasso \citep{Tibshirani94regressionshrinkage} is used to compute not only the R-vine structure, but also a regularization path of the R-vine matrix, allowing for flexible thresholding to reduce the model complexity based on a single structure. Even though the approach does not rely on maximum spanning trees, a $d$-dimensional R-vine needs to be fitted and stored, which is accompanied by the same computational complexity. Large computational gains are however obtained by setting a majority of all pair copulas to independence copulas upfront.\\
There have been attempts to relate undirected graphical models to R-vines for structure selection. \citet{Haffetal} showed that a $k$-truncated R-vine can be expressed as chordal graph with maximal clique size $k+1$ and vice versa. However, this chordal graph needs to adhere to some other non-trivial properties. These are in practice not met when a graphical model is fitted. Thus, finding a sparse undirected graph and translate it into a sparse R-vine is hard and computationally infeasible. \citet{KovacsSzantai2016} propose an algorithm to calculate a $k$-truncated R-vine from a chordal graph with maximal clique size $k+1$ and show that taking into account the aforementioned property only leads to a chordal graph with maximal clique size $k+2$. The later can then be used to estimate a $k+2$ truncated R-vine. However, the problem is only deferred. Where there exist several methods for finding sparse undirected graphical models in high dimensions, these will not be chordal in most cases. Finding a so called \textit{chordal cover} with clique size at most $k+1$ is known to be \textit{NP-complete}, see \citet{treewidth}. Hence, for dimensions $d > 1000$ where $k$ can not be assumed too small, the problem is also intractable.\\
Finally, there also exist Bayesian methods for R-vine estimation \citep{GruberCzado2015, GruberCzado20152}. However, these do require even more computational effort and are hence not feasible in more than $d > 20$ dimensions.

\subsection{Model Simplification}
All of the proposed approaches may be modified by testing each pair copula for the independence copula with density $c^\perp\left(u_1,u_2\right)=1$. Thus, a type-1 error $\alpha \in \left(0,1\right)$ is specified and each pair-copula is tested for the null hypothesis to be the independence copula. Only if this hypothesis can be rejected at the level $1-\alpha$, an estimation of pair copula family and parameter is performed. Similarly, entire R-vine trees can be tested for only containing the independence copula. However, this also requires that an additional tree is fitted before it can be tested. Contrary to this, a \textit{truncation level} $k \in \left\{1,\dots,d-2\right\}$ can be specified upfront. By doing so, only the first $k$ R-vine trees are estimated and independence is assumed for all higher trees. If a $k$-truncation is imposed, \eqref{eq:vinedensity} becomes
\begin{equation*}
f\left(x_1,\ldots,x_d\right) = \left(\prod_{i=1}^{d}~f_i\left(x_i\right)\right) \times \left(\prod_{i=1}^{k}~\prod_{e\in E_i}~c_{j\left(e\right),\ell\left(e\right);D\left(e\right)}\Big(F\left(x_{j\left(e\right)}|x_{D\left(e\right)}\right),F\left(x_{\ell\left(e\right)}|x_{D\left(e\right)}\right)\Big)\right).
\end{equation*}
In Example \ref{ex:exvine1}, a $k$-truncated R-vine is given by $c_{j,\ell|D} = c^\perp$ if $\left|D\right| \ge k$. More details can be found in \citet{brechmann-etal}. Finally, testing for independence does not explicitly decrease the computational effort. However, a $k$ truncation leads to only $kd$ pair copulas, whereas identifying a sensible truncation level prior to estimation is hard.

\section{Graphical Independence Models}\label{sec:undirectedgraphicalmodels}
To find structures for R-vines, we will use tools which have proven very successful in the Gaussian setting. More precisely, we will use models based on \textit{undirected graphs}.\\
Most of the terminology follows \citet[pp.\ 4--7]{Lauritzen1996}. Let $V \neq \emptyset$ be a finite set, the \textit{node set} and let $E \subseteq \left\{\left(\alpha,\beta\right)|\left(\alpha,\beta\right) \in V \times V \mbox{ with }\alpha \neq \beta\right\}$ be the \textit{edge set}. Thus, we obtain a graph $\GGG =\left(V, E\right)$ as pair of node set and edge set. We assume $\left(\alpha,\beta\right)\in E \Rightarrow \left(\beta,\alpha\right)\in E$, i.\,e.\ only \textit{undirected edges} and hence an \textit{undirected graph}. Define a \textit{path} of length $k$ from nodes $\alpha$ to $\beta$ by a sequence of distinct nodes $\alpha = \alpha_0, \ldots ,\alpha_k = \beta$ such that $\left(\alpha_{i-1},\alpha_i\right) \in E$ for $i=1,\ldots,k$. A \textit{cycle} is defined as a path with $\alpha=\beta$. A graph $\HHH=\left(W,\FFF\right)$ is a \textit{subgraph} of $\GGG = \left(V,E\right)$ if $W \subseteq V$ and $\FFF \subseteq E$. We have an \textit{induced subgraph} $\HHH=\left(W,\FFF\right)$ if $W \subseteq V$ and $\FFF=\left\{\left(\alpha,\beta\right)|\left(\alpha,\beta\right) \in W \times W \mbox{ with } \alpha \neq \beta\right\} \cap E$, i.\,e.\ $\HHH$ contains a subset of nodes of $\GGG$ and all the edges existing between these nodes in $\GGG$. If a path from $\alpha$ to $\beta$ exists for all $\alpha,\beta \in V$, we say that $\GGG$ is \textit{connected}. Whenever for a graph $\GGG = \left(V,E\right)$ we have that there exists a disjoint partition of $V = \bigcup_{i=1}^p~V_i$ such that the $p$ subgraphs $\HHH_i$ induced by $V_i$ for $i=1,\dots,p$ are connected subgraphs, we speak of \textit{connected components} of $\GGG$. For $\GGG$ undirected, $\alpha,\beta \in V$, a set $S \subseteq V$ is said to be an $\left(\alpha,\beta\right)$ \textit{separator} in $\GGG$ if all paths from $\alpha$ to $\beta$ intersect $S$. $S$ is said to \textit{separate} $A$ from $A$ in $\GGG$ if it is an $\left(\alpha,\beta\right)$ separator in $\GGG$ for every $\alpha \in A$, $\beta \in B$ and we denote it by $\separate{\alpha}{\beta}{S}{\GGG}$ and $\separate{A}{B}{S}{\GGG}$, respectively. We define the \textit{adjacency matrix} $\varPi^\GGG = \left(\pi^\GGG_{i,j}\right)_{i,j=1,\dots,d} \in \left\{0,1\right\}{d \times d}$ of a graph $\GGG=\left(V,E\right)$ such that $\pi^ \GGG_{i,j} = \pi^\GGG_{j,i} = 1 \Leftrightarrow \left(i,j\right) \in E$ and $0$ otherwise.

\subsection{Probabilistic (Gaussian) Graphical Models}
Graph theory and statistical models can be linked, obtaining \textit{probabilistic graphical models}. More precisely, consider a random vector $\bm{X} = \left(X_1,\dots,X_d\right)$ and assume $\bm{X} \sim \NNN_d\left(\mathbf{0},\Sigma\right)$, denoting a $d$-dimensional Gaussian distribution with probability density function
\begin{equation}\label{eq:mvGaussiandensity}
f\left(\bm{x};\bm{0},\Sigma\right)=\left(2\pi\right)^{-\frac{p}{2}} \det\left(\Sigma\right)^{-\frac{1}{2}}\exp\bigg(-\frac{1}{2}\bm{x}^T\Sigma^{-1}\bm{x}\bigg).
\end{equation}
Let $i,j \in \left\{1,\dots,d\right\}$ with $i \neq j$ and $S \subseteq \left\{1,\dots,d\right\}\setminus\left\{i,j\right\}$. We denote that $X_i$ is conditionally independent of $X_j$ given $\bm{X}_S$ by $\condindep{X_i}{X_j}{\bm{X}_S}$. In the remaining, let $\GGG=\left(V=\left\{1,\dots,d\right\}, E\right)$. We say that $\bm{X}$ is \textit{Markov} with respect to $\GGG$ when
\begin{equation}\label{eq:markov}
\condindep{X_j}{X_\ell}{\left\{X_1,\dots,X_d\right\}\setminus\left\{X_j,X_\ell\right\}} \Leftrightarrow \left(j,\ell\right) \notin E.
\end{equation}
This means, each missing edge in the graph $\GGG$ corresponds to random variables which are conditionally independent given the rest. The later is however also directly expressed via the inverse of the correlation matrix. More precisely, let $\Omega = \Sigma^{-1}$, then
\begin{equation*}
\Omega_{j\ell} = 0 \Leftrightarrow \condindep{X_j}{X_\ell}{\left\{X_1,\dots,X_d\right\}\setminus\left\{X_j,X_\ell\right\}}.
\end{equation*}
Thus, estimating the \textit{sparsity pattern} of $\Omega$ is equivalent to estimating the graph $\GGG$. From the graph however, we can also extract more information. If the underlying distribution is Gaussian, we consider the \textit{global Markov property} such that for $S \subseteq \left\{1,\dots,d\right\}\setminus\left\{i,j\right\}$ we have that
\begin{equation}\label{eq:globalmarkov}
\condindep{X_j}{X_\ell}{\bm{X}_S} \Leftrightarrow \separate{j}{\ell}{S}{\GGG},
\end{equation}
which is a more favourable property than \eqref{eq:markov} and which we will exploit later. Next, we will focus on how to estimate such graphs, i.\,e.\ the matrix $\Omega$, given data.
\begin{example}[Graphical model]\label{ex:graphicalmodel}
Consider the graph in Figure \ref{fig:graph}.
\begin{figure}[h]
	\centering
	\includegraphics[width=0.25\textwidth]{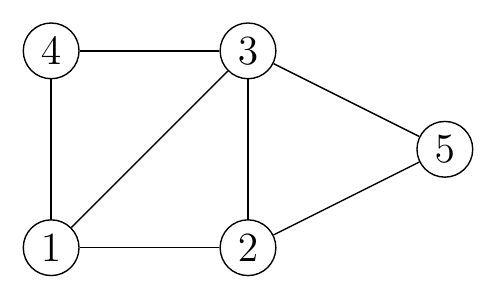}
	\caption{Graphical model in $5$ dimensions.}
	\label{fig:graph}
\end{figure}
By virtue of the graphical separation, we have e.\,g.\ $\separate{1,4}{5}{2,3}{\GGG} \Rightarrow \condindep{1,4}{5}{2,3}$.
\end{example}

\subsection{Estimating Sparse Inverse Covariance Matrices with the Graphical Lasso}
To estimate a sparse precision matrix $\Omega$, we use the well-known \textit{graphical Lasso}, see \citet{glasso}. Denote the sample covariance matrix by $S = X^TX/n \in \mathbb{R}^{d \times d}$ where $X = \left(x_{i,j}\right)_{i=1,\dots,n,j=1,\dots,d} \in \mathbb{R}^{n \times d}$ is the observed and centred data matrix with $\bm{X} \sim \NNN_d\left(\bm{0},\Sigma\right)$. Then, the graphical Lasso calculates a sparse undirected graphical model by finding a solution for $\Omega$. Considering the logarithm of \eqref{eq:mvGaussiandensity} and taking derivatives with respect to $\Sigma$ we obtain as optimization problem
\begin{equation}\label{eq:loglik}
\max_{\Omega \in \mathbb{R}^{d \times d}}~\log\left(\det\left(\Omega\right)\right) + \tr\left(S\Omega\right).
\end{equation}
with solution $\widehat{\Omega} = S^{-1}$. However, this will in general have no zero entries and hence, induce no sparse graph. Furthermore, in high dimensional data sets we often have  $d > n$, which leads to a singular matrix $S$ and the inverse of $S$ does not exist. The graphical Lasso overcomes this by introducing a penalty in \eqref{eq:loglik} and solving 
\begin{equation}\label{eq:glassologlik}
\max_{\Omega \in \mathbb{R}^{d \times d}}~\log\left(\det\left(\Omega\right)\right) + \tr\left(S\Omega\right) + \lambda \sum_{i=1}^d~\sum_{j=1}^d~\left|\Omega_{ij}\right|.
\end{equation}
depending on some \textit{regularization parameter} $\lambda \geq 0$. For this optimization problem, many efficient numerical solvers also for thousands of dimensions exist. For $\lambda = 0$, there is no penalization and the solutions of \eqref{eq:loglik} and \eqref{eq:glassologlik} coincide. For fixed $\lambda > 0$, denote the solution of \eqref{eq:glassologlik} by $\widehat{\Omega}^\lambda$ and define $\GGG^\lambda = \left(V,E^\lambda\right)$ by
\begin{equation}\label{eq:glassograph}
(i,j) \in E^\lambda \Leftrightarrow \widehat{\Omega}^\lambda_{ij} \neq 0.
\end{equation}
Varying $\lambda > 0$, we obtain a piecewise constant \textit{solution path} of graphs with different levels of sparsity. Letting $\lambda \to 0$, the solutions $\GGG^\lambda$ will become more and more dense. For $\lambda \to \infty$, the number of \textit{connected components} will increase but their individual sizes decreases. In practice, the solution path is calculated along a vector $\bm{\lambda}=\left(\lambda_1,\dots,\lambda_J\right)$ with $\lambda_j > 0$. Several modifications and improvements for the graphical Lasso have been proposed. For instance, \citet{meinshausen2006} show that neighbourhood selection consistently estimates the graph $\GGG$, however, no estimate $\widehat{\Omega}$ is obtained. \citet{glasso2} demonstrate that the search for \eqref{eq:glassologlik} can be carried out in terms of block-diagonal matrices, breaking apart the large problem into smaller ones.\\
Vital for the first part of our proposed approach is that the connected components with respect to some $\lambda > 0$ can also be calculated directly from the sample covariance matrix $S$. More precisely, consider a fixed $\lambda > 0$ and a solution
$\widehat{\Omega}^\lambda$ of \eqref{eq:glassologlik} with $\GGG^\lambda = \left(V,E^\lambda\right)$ as defined in \eqref{eq:glassograph}. Thus, $\widehat{\Omega}^\lambda$ and equivalently $E^\lambda$ induce a vertex partition
\begin{equation*}
V = \bigcup_{i=1}^p~V^\lambda_i,
\end{equation*}
where each $V^\lambda_i$ is a connected component for $i=1,\dots,p$, based on the edge set $E^\lambda$.\\
Alternatively, define a graph $\HHH = \left(V, \FFF^\lambda\right)$ based on the sample covariance matrix $S$ with edge set $\FFF^\lambda$ and adjacency matrix $\varPi^\HHH$ such that
\begin{equation}\label{eq:screening}
\pi^\HHH_{i,j} = 1 \Leftrightarrow \left(i,j\right) \in \FFF^\lambda \Leftrightarrow \left|S_{ij}\right| \ge \lambda,
\end{equation}
This way of assigning the edge set is called \textit{screening}. The graph $\HHH$ has now, say $q$ connected components we denote by $W_i$ for $i=1,\dots,q$ and consider the associated partition $V = \bigcup_{i=1}^q~W_i$. It has now been shown by \citet{glasso3}, that $p=q$ and moreover, $V_i = W_i$ for all $i=1,\dots,p$. This makes a decomposition of the entire graphical Lasso problem more tractable as we can split it into $p$ parallel tasks which can be performed entirely independent from each other. Furthermore, we have a very easy screening rule for intractably high dimensional datasets to decompose their dependence behaviour in multiple smaller parts which now are tractable. Inside the connected components, we can then use the graphical Lasso to obtain non-dense graphs. We give a brief example.
\begin{example}[Screening]\label{ex:clustering:1}
Assume we have a dataset on the \textit{z-scale} in $6$ dimensions with the following empirical covariance matrix
\begin{equation*}
S =
	\left(
	\begin{array}{rrrrrr}
		1.0000 & 0.2058 & 0.1794 & 0.7340 & 0.7298 & 0.7167 \\ 
		0.2058 & 1.0000 & 0.3212 & 0.2643 & 0.3158 & 0.2848 \\ 
		0.1794 & 0.3212 & 1.0000 & 0.1895 & 0.2105 & 0.2327 \\ 
		0.7340 & 0.2643 & 0.1895 & 1.0000 & 0.9606 & 0.9089 \\ 
		0.7298 & 0.3158 & 0.2105 & 0.9606 & 1.0000 & 0.9378 \\ 
		0.7167 & 0.2848 & 0.2327 & 0.9089 & 0.9378 & 1.0000 \\ 
	\end{array}
	\right).
\end{equation*}
We use the \texttt{huge} R-package \citep{hugemanual} to calculate a sequence of $J=4$ values for $\lambda$ given by
\begin{equation*}
\bm{\lambda} = \left(0.9607, 0.7438, 0.3452, 0.2070\right).
\end{equation*}
Solving \eqref{eq:glassologlik} for these values we obtain graphical models $\GGG^{\lambda_1},\dots,\GGG_{\lambda_4}$ as shown in Figure \ref{fig:ex:clustering1}.
\begin{figure}[h]
	\centering
	\begin{tabular}{c||c}
		$\GGG^{\lambda_1}$, $\lambda_1 = 0.9607$& 		$\GGG^{\lambda_2}$, $\lambda_2 = 0.7438$\\
		\hline
	\includegraphics[width=0.3\textwidth, trim={0.1cm 0.1cm 0.1cm 0.1cm},clip]{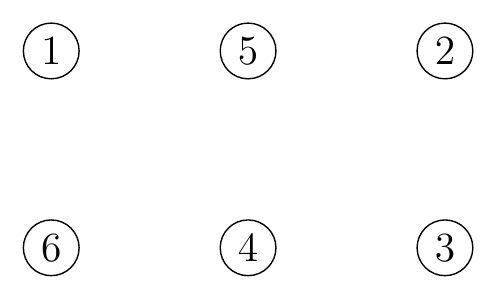}&
	\includegraphics[width=0.3\textwidth, trim={0.1cm 0.1cm 0.1cm 0.1cm},clip]{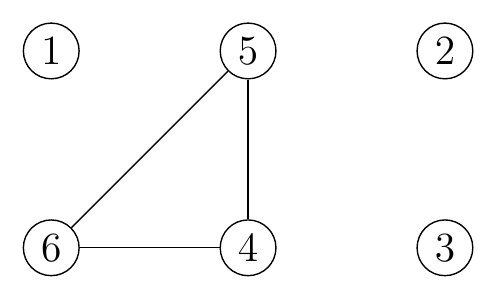}\\
	\hline \hline
			$\GGG^{\lambda_3}$, $\lambda_3 = 0.3452$& 		$\GGG^{\lambda_2}$, $\lambda_4 = 0.2070$\\
	\hline
	\includegraphics[width=0.3\textwidth, trim={0.1cm 0.1cm 0.1cm 0.1cm},clip]{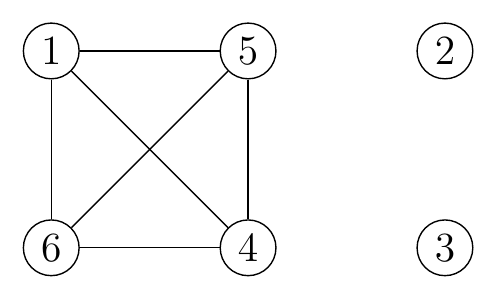}&
	\includegraphics[width=0.3\textwidth, trim={0.1cm 0.1cm 0.1cm 0.1cm},clip]{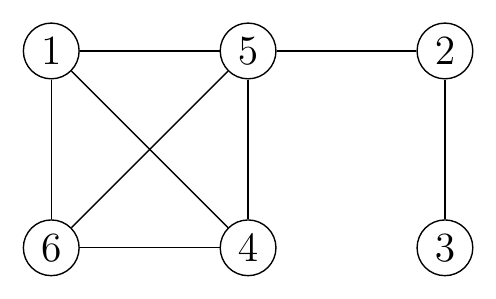}
	\end{tabular}
	\caption{Example \ref{ex:clustering:1}: Sequence of estimated graphical models $\GGG^{\lambda_1},\dots,\GGG^{\lambda_4}$.}
	\label{fig:ex:clustering1}
\end{figure}
Initially, there are only isolated nodes and hence $\left|V\right| = 6$ connected components. In the second and third graph, we have one connected component of size $3$ together with isolated nodes and size $4$ with isolated nodes. In the fourth graph, only one connected components of size $\left|V\right| = 6$ exists. If we consider $\lambda_2 = 0.7438$ and apply the screening rule \eqref{eq:screening}, we obtain the following adjacency matrix based on $S$
\begin{equation*}
\varPi^{\GGG^{\lambda_2}} = \left(
\begin{array}{rrrrrr}
	 0 & 0 & 0 & 0 & 0 & 0 \\ 
	 0 & 0 & 0 & 0 & 0 & 0 \\
	 0 & 0 & 0 & 0 & 0 & 0 \\ 
	 0 & 0 & 0 & 0 & 1 & 1 \\ 
	 0 & 0 & 0 & 1 & 0 & 1 \\ 
	 0 & 0 & 0 & 1 & 1 & 0 \\ 
\end{array}
\right).
\end{equation*}
which constitutes the same connected components as in the corresponding graph of the graphical Lasso solution path. The adjacency matrix $\varPi$ indicates that the nodes $4,5,6$ are all connected to each other. However, the result by \citet{glasso3} only assures that the connected components of the screening rule and the graphical Lasso are the same for a specific $\lambda > 0$, but no knowledge about the edges in the connected components is can be drawn from the screening rule.
\end{example}

\section{Selection of High Dimensional R-vines with Graphical Models}\label{sec:selectvines}
Recall that our goal is ultimately to estimate R-vines in ultra high dimensions. Thus instead of estimating an R-vine as described in Section \ref{sec:rvines} on data in $d$ dimensions, we first cluster the dataset into a partition and then perform estimation on the elements of the partition. The partition is provided by the methods presented in Section \ref{sec:undirectedgraphicalmodels}, i.\,e.\ the graphical Lasso. Inside these clusters, we can then estimate R-vines in smaller dimensions with improved accuracy as a consequence of exploiting the global Markov property, \eqref{eq:globalmarkov}.
\subsection{Clustering High Dimensional Data}\label{subsec:clustering}
Assume that we are given data $X \in \mathbb{R}^{n \times d}$, where $d \approx 1000$ or more. We will consider a sequence of $J$ disjoint partitions of $V = \left\{1,\dots,d\right\}$ into $p_j$ connected components $V^j \coloneqq \bigcup_{i=1}^{p_j}~V_i^j$, $j=1,\dots,J$. This is expressed by a sequence of graphical models
\begin{equation}\label{eq:graphseq}
\GGG_1=\left(V^1=\bigcup_{i=1}^{p_1}~V_i^1,\bigcup_{i=1}^{p_1}~{E}_i^1\right),\dots,\GGG_J=\left(V^J=\bigcup_{i=1}^{p_J}~V_i^J,\bigcup_{i=1}^{p_J}~{E}_i^J\right).
\end{equation}
In most practical applications, $J=15$ or $J=30$, see \citet{hugemanual}. If partition $V^j$ is only a single connected component, we have $p_j=1$. The sequence $\GGG_j$ for $j=1,\dots,J$ can be identified as solution path corresponding to the graphical Lasso for $J$ different penalization values of $\lambda > 0$. To identify the size of connected components in these graphs, define for each partition $V^j$, $j=1,\dots,J$,
\begin{equation}\label{eq:delta}
\delta_j = \max_{i = 1,\dots,p_j}~\left|V_i^j\right|.
\end{equation}
Instead of considering the entire dataset in $d$ dimensions, we consider subsets of lower dimensions on the connected component with maximum dimension $\delta_j < d$ dimensions. In practical applications, we will have some \textit{threshold dimension} $0 < d_T < d$ and calculate the solution path of the graphical Lasso for a sequence $\left\{\lambda_1,\dots,\lambda_J\right\}$ based on the screening property \eqref{eq:screening}. This works very fast and we can select the corresponding graphical model and associated partition $V^T$ by
\begin{equation}\label{eq:thresholddim}
T = \argmax_{j=1,\dots,J}~\delta_j \mbox{ such that } \delta_j \leq d_T.
\end{equation}
Finally, we denote the chosen partition $T$ by $V^{T}$ and the corresponding graph by $\GGG_T=\left(V^T=\bigcup_{i=1}^{p_{T}}~V_i^T,\bigcup_{i=1}^{p_{T}}~{E}_i^T\right)$. 
\begin{example}[Example \ref{ex:clustering:1} cont.]\label{ex:clustering:2}
	Consider the sequence of graphs in Example \ref{ex:clustering:1}. Using the notation of \eqref{eq:graphseq} and defining $\GGG_j = \GGG^{\lambda_j}$ , we have $J=4$ and $p_j, \delta_j$ for $j=1,\dots,4$ as follows:
	\begin{table}[h]
		\centering
		\begin{tabular}{c||cccc}
			$j$	& 1 & 2 & 3 & 4 \\
			\hline\hline
			$p_j$ & 6 & 4 & 3 & 1 \\
			\hline
			$\delta_j$ & 1 & 3 & 4 & 6
		\end{tabular}
		\caption{Example \ref{ex:clustering:2}: Number of connected components $p_j$ and maximal component sizes $\delta_j$ for graphs in Figure \ref{fig:ex:clustering1}.}
		\label{tab:ex:clustering}
	\end{table}
	Assume $d_T = 4$, then $T = \argmax_{j=1,\dots,4}~\delta_j \mbox{ such that } \delta_j \leq 4$, thus $T=3$. Hence, the graph $\GGG_3$ is selected with partition $V^3 = \left\{1,4,5,6\right\} \cup \left\{2\right\} \cup \left\{3\right\}$ with $p_T = 3$ and $\delta_T = 4$.
\end{example}
Now, we consider the problem of estimating \textit{sub-R-vines} on these connected components induced by a partition $V^T$. Thus, with respect to this partition we estimate R-vines on the connected components of $\GGG_T$, i.\,e.\ the elements of the partition $V^T_i$, $i=1,\dots,p_T$ with at most dimension $\delta_T$. Each of these \textit{sub-R-vines} is then stored in an R-vine matrix and the corresponding matrices can be combined non-uniquely to an R-vine matrix of dimension $d \times d$. This is however an incomplete R-vine matrix as it does not contain information how the connected components are connected to each other. Additionally, connected components of size $1$, i.\,e.\ isolated nodes are not yet included. Both the missing connections and the isolated nodes can however be easily connected afterwards as we will show in a subsequent example. For this, we also introduce the \textit{fill-level} $k_F \geq 0$ which determines how many R-vine trees \textit{outside} the connected components should be estimated.
\begin{example}[Toy example]\label{ex:toy}
	Assume we have the graphical model $\GGG_T$ as in Figure \ref{fig:ex:toy}.
	\begin{figure}[h]
		\centering
		\includegraphics[width=0.25\textwidth, trim={0.1cm 0.1cm 0.1cm 0.1cm},clip]{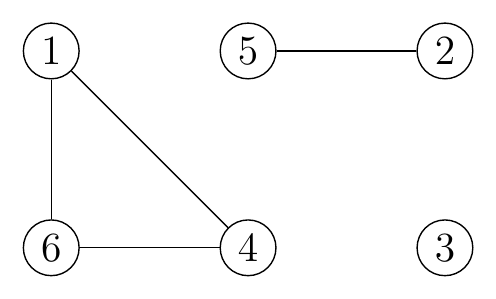}
		\caption{Example \ref{ex:toy}: Graphical model $\GGG_T$ in $6$ dimensions.}
		\label{fig:ex:toy}
	\end{figure}
	Given this graphical model, we estimate \textit{sub-R-Vines} on the components $\left\{1,4,6\right\}$ and $\left\{2,5\right\}$, respectively. These two components give rise to the following R-vine matrices, where we assume that the estimate in $M_1$ is optimal with respect to some edge weight.
	\begin{equation*}
	\begin{aligned}
	M_1 = \left(
	\begin{array}{ccc}
	1 & & \\
	6 & 4 & \\
	4 & 6 & 6
	\end{array}
	\right)
	&\vspace{2cm}&
	M_2 = \left(
	\begin{array}{cc}
	2 & \\
	5 & 5
	\end{array}
	\right).
	\end{aligned}
	\end{equation*}
	Together with the isolated node $3$, these two R-vines can be arranged into a \textit{joint} R-vine on $6$ dimensions described by the matrix $M_{123}$.
	\begin{equation*}
	M_{123} = \left(
	\begin{array}{cccccc}
	3 & & & & & \\
	& 2 & & & & \\
	& & 5 & & & \\
	& & & 1 & & \\
	\hline
	& & & 6 & 4 & \\
	\Box_1 & 5 & \Box_2 & 4 & 6 & 6
	\end{array}
	\right),
	M_{123}' = \left(
	\begin{array}{cccccc}
	3 & & & & & \\
	& 2 & & & & \\
	& & 5 & & & \\
	& & & 1 & & \\
	\hline
	\triangle_1 & \triangle_2 & \triangle_3 & 6 & 4 & \\
	2 & 5 & 1 & 4 & 6 & 6
	\end{array}
	\right).
	\end{equation*}
	This would correspond to the graphical model in $\GGG_T$. Now, we can connect the connected components with each other in the first $k_F$ trees, i.\,e.\ to the fill-level. This works as following for the example of $k_F = 2$.
	The entries in $M_{123}$ marked by $\Box$ describe the pair copulas between the connected components and are chosen from $\Box_1 \in \left\{2,5,1,4,6\right\}$ and $\Box_2 \in \left\{1,4,6\right\}$. For example, $\Box_2 \notin \left\{3,2\right\}$ since only diagonal entries from the right of the corresponding column may be used. To select the entries in the last row, we consider all admissible pairs $\left(3,\Box_1\right)$ with $\Box_1 \in \left\{2,5,1,4,6\right\}$ and $\left(5,\Box_1\right)$ with $\Box_2 \in \left\{1,4,6\right\}$. Recall that a pair copula in the first tree is fitted on the pair of diagonal entry and the entry in the last row of the R-vine matrix. We fit pair copulas for each of these pairs and then select the best according to some edge weight $\mu$.	
	After having completed the last row in $M_{123}$ using $\Box_1 = 2$ and $\Box_2 = 1$, we obtain $M_{123}'$ and are to fill the second tree, i.\,e.\ the fifth row of $M_{123}'$ consisting of the entries $\triangle_1,\triangle_2,\triangle_3$.
	However, we have to take into account the proximity condition. By checking \eqref{eq:proximitycondition3}, this leaves admissible entries $\triangle_1 \in \left\{5\right\}$, $\triangle_2 \in \left\{1\right\}$ and $\triangle_3 \in \left\{4\right\}$. Finally, these pair copulas are fitted and the matrix is finalized, see $M_{123}''$ with associated family matrix $\Gamma$ with $\star$ denoting pair copulas which are not the independence copula by virtue of the graphical model.
	\begin{equation*}
	M_{123}'' = \left(
	\begin{array}{cccccc}
	3 & & & & & \\
	6& 2 & & & & \\
	4& 6& 5 & & & \\
	1& 4& 6& 1 & & \\
	\hline
	5 & 1 & 4 & 6 & 4 & \\
	2 & 5 & 1 & 4 & 6 & 6
	\end{array}
	\right),
	\Gamma = \left(
	\begin{array}{cccccc}
	 & & & & & \\
	&  & & & & \\
	& &  & & & \\
	& & &  & & \\
	\hline
	\star & \star & \star & \star & \\
	\star & \star & \star & \star & \star & 
	\end{array}
	\right).
\end{equation*}
	We note that this particular R-vine is a \textit{D-vine}, i.\,e.\ the first R-vine tree is given by a path through the nodes $3-2-5-1-4-6$ and determines all subsequent trees. This is however not necessarily the case in general.
\end{example}
Motivated by the previous example, we define the \textit{R-vine representation} of an undirected graphical model $\GGG$ with fill level $k_F \geq 0$ by $\VVV\left(\GGG,k_F\right)$.\\
Thus, we are left to estimate R-vines in the connected components of $\GGG_T$ which can be combined into one R-vine. We will not use the standard algorithm for estimation but exploit also the graphical structure \textit{within the connected component}. Hence, consider an arbitrary connected component within $\GGG_T$ with of size $\nu$ and denote it by $\HHH=\left(W,\FFF\right)$. This is again a graphical model with respect to the vertices in $W$. We will describe a very efficient approach for estimating R-vines with improved accuracy.

\subsection{Improving Estimation Accuracy}\label{subsec:improvingaccuracy}
We are now considering a connected component of $\GGG_T = \left(V^T=\bigcup_{i=1}^{p_{T}}~V_i^T,\bigcup_{i=1}^{p_{T}}~{E}_i^T\right)$ and denote it by $\HHH = \left(W, \FFF\right)$. We are to estimate an R-vine on the variables in $W$, denote $\nu = \left|W\right|$. We consider the computational complexity in terms of pair copulas to estimate. In total, these are $\nu\left(\nu-1\right)/2$ pair copulas to be estimated for a R-vine on $W$ with $\nu$ variables. Denote the corresponding R-vine tree sequence $\VVV = \left(T_1,\dots,T_{d-1}\right)$. To find the first R-vine tree $T_1$, start with a full graph on $W$. Di{\ss}mann's algorithm would now Kendall's $\tau$ on all pairs $\left(j,\ell\right) \in W \times W$ and use the weights $\mu_{j,\ell} = \left|\tau_{j,\ell}\right|$ to find a \textit{maximum spanning tree}. As discussed in Section \ref{subsec:modelselection}, we can also estimate pair copula densities $c_{j,\ell}$ for each edge $\left(j,\ell\right)$ on all $\nu\left(\nu-1\right)/2$ edges and calculate a maximum spanning tree with respect to an edge weight $\mu_{j,\ell}$, e.\,g.\ log-likelihood or AIC, based on the \textit{actual fit}. However, this increases the required effort for an R-vine tree sequence significantly. As we have $\nu-1$ trees on $\nu-i+1$ nodes for $i=2,\dots,\nu-2$ and consider in the worst case all possible pairs in each tree, this sums up to 
\begin{equation}\label{eq:complexity}
\sum_{i=1}^{\nu-2}~\frac{\left(\nu-i+1\right)\left(\nu-i\right)}{2} = \frac{\nu^3-\nu}{6} - 1,
\end{equation}
pair copulas, where the leading term can only be bounded from above by order $\nu^3$. Note that this is a worst case complexity since in higher trees, the proximity condition can exclude some edges, and hence, no pair copula needs to be estimated for these. In the particular case of a \textit{D-vine}, where the first tree is a path, i.\,e.\ each node except for the two end nodes has degree $2$, the remaining R-vine tree sequence is completely determined and the effort collapses to order $\nu\left(\nu-1\right)/2$. Whereas in the case of a \textit{C-vine}, where each tree is a star, the upper bound in \eqref{eq:complexity} is attained. Recall that the method of \citet{dissmann-etal}, only $\nu\left(\nu-1\right)/2$ pair copulas have to be estimated. Hence, for general R-vines, estimation of $\nu^3$ pair copulas is not admissible even if $\nu \ll d$. However, we can take into account the (conditional) independence information by the graph $\HHH$ to set a huge number of pair copulas to independence copulas upfront, leading to a significant decrease of computational effort. 
\begin{itemize}
	\item Recall that $\FFF$ is the edge set of $\HHH$ and define $\HHH_0 = \left(W, W \times W\right)$, i.\,e.\ a full graph on $W$ and assign pair copulas and weights
	\begin{equation}\label{eq:vineesttree1}
	c_{j,\ell} = 1 \Leftrightarrow \mu_{j,\ell} = 0 \Leftrightarrow \left(i,j\right) \notin \FFF.
	\end{equation}
	For all remaining pairs $\left(j,\ell\right) \in W \times W$, we perform maximum likelihood estimation on the pair copulas and obtain weights $\mu_{j,\ell}$ based on the actual fit. Thus, we have weights for all edges and can now calculate a maximum spanning tree $T_1 = \left(W, E_1\right)$.
	\item We define $T_k = \left(E_{k-1}, E_{k-1} \times E_{k-1}\right)$ for $k=2,\dots,\nu-1$, i.\,e.\ a full graph on $E_{k-1}$. We remove all edges not allowed by the proximity condition. For the remaining edges, we assign pair copulas and weights 
	\begin{equation}\label{eq:vineesttreehigher}
	c_{j,\ell|\mathbf{D}} = 1 \Leftrightarrow \mu_{j,\ell|\mathbf{D}} = 0 \Leftrightarrow \separate{j}{\ell}{\mathbf{D}}{\HHH}.
	\end{equation}
	For all remaining pairs $\left(j,\ell\right) \in W \times W$ we perform maximum likelihood estimation on the pair copulas and obtain weights $\mu_{j,\ell}$ based on the actual fit. Thus, we have weights for all edges and can now calculate a maximum spanning tree $T_k$ for $k = 2,\dots,\nu-1$.
\end{itemize}
If the connected component $\HHH$ is not too dense, i.\,e.\ $\left|\FFF\right| \sim \nu$ instead of $\left|\FFF\right| \sim \nu^2$, overall about $\nu^2$ pair copulas are estimated for the entire R-vine in $\nu$ dimensions. From our point of view, it is more beneficial to neglect the edges which are not chosen anyway because of (conditional) independence and perform a more thorough analysis on the remaining edges. The algorithm of Di{\ss}mann might miss important edges, especially if all estimated Kendall's $\tau$ values are similar. Furthermore, our approach fosters sparsity by setting pair copulas to independence straight away. This is not possible in Di{\ss}mann's algorithm, except for truncation, which however has to be specified upfront in a very inflexible manner. We will show that our approach is able to capture significantly more dependence compared to Di{\ss}mann's. An example illustrating this approach finalizes the section.
\begin{example}[Example \ref{ex:clustering:2} cont.]\label{ex:clustering:3}
Assume the graphical model $\HHH$ as in Figure \ref{fig:ex:clustering:3:1} (left).
\begin{figure}[h]
	\centering
	\includegraphics[width=0.175\textwidth, trim={0.1cm 0.1cm 0.1cm 0.1cm},clip]{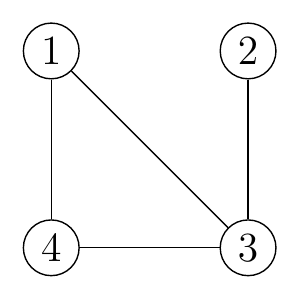}
	\includegraphics[width=0.175\textwidth, trim={0.1cm 0.1cm 0.1cm 0.1cm},clip]{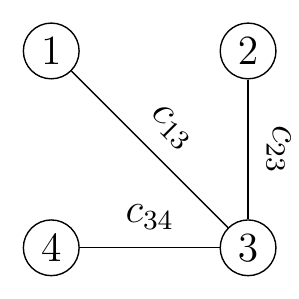}
	\caption{Example \ref{ex:clustering:3}: Graphical model $\HHH$ (left) and first R-vine tree $T_1$ (right).}
	\label{fig:ex:clustering:3:1}
\end{figure}
Using our approach, we fit pair copulas $c_{13}, c_{14}, c_{23}, c_{34}$ and consider the corresponding goodness of fit values $\mu_{13},\mu_{14},\mu_{23},\mu_{34}$, given for example by the $\AIC$. The other missing edges are neglected, i.\,e.\ no pair copula is estimated. Assume additionally that the optimal choice with respect to AIC is given by the tree given in the right panel of Figure \ref{fig:ex:clustering:3:1}. Then, the second R-vine tree $T_2$ may contain for example the possible edge $c_{24|3}$. However, we can see from the left panel of Figure \ref{fig:ex:clustering:3:1} that $\separate{2}{4}{3}{\HHH}$ and thus, the pair copula $c_{24|3}$ is set to the independence copula upfront in $T_2$.
\end{example}

\subsection{Implementation of the Algorithm}
We combine the two previous steps in one algorithm, allowing to estimate high dimensional vine copulas based on a clustering by the graphical Lasso. For this, we need to specify a threshold dimension $d_T < d$ and a fill-level $k_F \geq 0$ describing until which tree we will estimate pair-copulas \textit{outside} the connected components. The later is beneficial since we only use the connected components to break our original problem into tractable sub-problems. However, we assume there is dependence outside the connected components. From a computational point of view, estimating within the connected components of dimension at most $d_T < d$ first and then connect these components afterwards is much more beneficial than estimating an R-vine on $d$ dimensions. We use the \texttt{huge} R-package, see \citet{hugemanual} to generate high dimensional undirected graphical models and use the default settings there. Since we normally operate on \textit{copula-data} $\left(U_1,\dots,U_n\right)$, i.\,e.\ data with uniform marginals, we transform our observations to the \textit{z-scale}, i.\,e.\ consider $\left(Z_1 = \Phi^{-1}\left(U_1\right),\dots,Z_d = \Phi^{-1}\left(U_d\right)\right) \sim \NNN\left(0,\Sigma\right)$ in a Gaussian set up. The only change we perform in the default setting of \texttt{huge} is that we always want to obtain a sequence of $30$ graphs, this is regulated by the number of $\lambda_j \geq 0$ to be evaluated. In terms of the previous notation, we set $J=30$. Our algorithm then selects a partition such that the maximal component size is less or equal $d_T$ as shown in Section \ref{subsec:clustering} and then performs on each of the components an improved R-vine selection based on Section \ref{subsec:improvingaccuracy}. The edge metric $\mu$ we use is the AIC of the associated pair copula term in our considerations. Afterwards, we combine these \textit{sub-R-vines} into one \textit{joint} R-vine matrix on which we operate further. Finally, we estimate the pair copulas in the first $k_F \geq 0$ trees of the joint R-vine. From this point on, we only operate on the R-vine matrix $M$. Thus, we have to take into account that in a certain column $j$ of $M$, only values of the main diagonal right of $j$, i.\,e.\ in $M_{j+1,j+1},\dots,M_{d,d}$ can occur. Thus, for a given entry in $M_{i,j}$, we check which entries in $M_{j+1,j+1},\dots,M_{d,d}$ are valid according to the proximity condition. For those, we fit pair copulas and select the best choice according to some metric $\mu$, e.\,g.\ $\mu = \AIC$. The entire algorithm, which we will refer to as \textit{RVineClusterSelect}, is given Appendix, Section \ref{sec:appendix:algorithm}.

\section{Numerical Examples}\label{sec:numericalexamples}
To show the feasibility of our approach, we present different numerical examples.
\begin{enumerate}
	\item A simulation study of several sparse dependence models with respect to a subset in $d=85$ dimensions of the S\&P100 index constituents. We compare our approach to the Di{\ss}mann algorithm with respect to goodness of fit and computation time.
	\item A runtime comparison with Di{\ss}mann's in up to $d=1750$ dimensions.
	\item An ultra high-dimensional data application involving $d = 2131$ stocks from different sectors and multiple geographies. We compare our approach to Gaussian models.
\end{enumerate} 

\subsection{Preprocessing of Data}
Because of the wide availability, we apply our methods to high dimensional financial daily data. More precisely, we consider closing prices of shares adjusted for dividends and splits $S_i^j$ with $i = 1,\dots,n$ observations for $j = 1,\dots,d$ shares. In the next step, we calculate daily log returns by $R_i^j = \log\left(S_i^j/S_{i-1}^j\right)$ for $i = 2,\dots,n$. These log returns are then filtered for trend and seasonality, i.\,e.\ idiosyncratic behaviour. We use ARMA-GARCH$\left(p,q\right)$ models with $\left(p,q\right) \in \left\{0,1\right\} \times \left\{0,1\right\}$, i.\,e.\ four different specifications and allow for residuals distributed according to three different distributions, Gaussian, Student's-t or skewed Student's-t. In total we consider $4 \times 3=12$ models which we fit for each marginal time series and choose the best in terms of log-likelihood. Next, we compute standardized residuals $x_{ij} = \left(R_i^j - \widehat{R}_i^j\right)/\widehat{\sigma}_{j}$ for $j=1,\dots,d$ and $i=1,\dots,n$ based on the estimated time series models, where $\widehat{\sigma}^2_{j}$, $j=1,\dots,d$ is the estimated variance of the error distribution. Finally, we calculate the \textit{empirical cumulative distribution function} $\widehat{F}_j$ of $x_{1j},\dots,x_{nj}$ for $j=1,\dots,d$ to obtain \textit{copula data} $U_{ij} = \widehat{F}_j\left(x_{ij}\right)$ for $i=1,\dots,n$ and $j=1,\dots,d$, on which we work on. The data was acquired using the \texttt{quantmod} R-package \citep{quantmod} and the marginal time series models were fitted using the \texttt{rugarch} R-package \citep{rugarch}.

\subsection{Simulation Study}
\begin{figure}[h]
	\centering
	\includegraphics[width=0.48\textwidth, trim={0.1cm 0.1cm 0.1cm 0.1cm},clip]{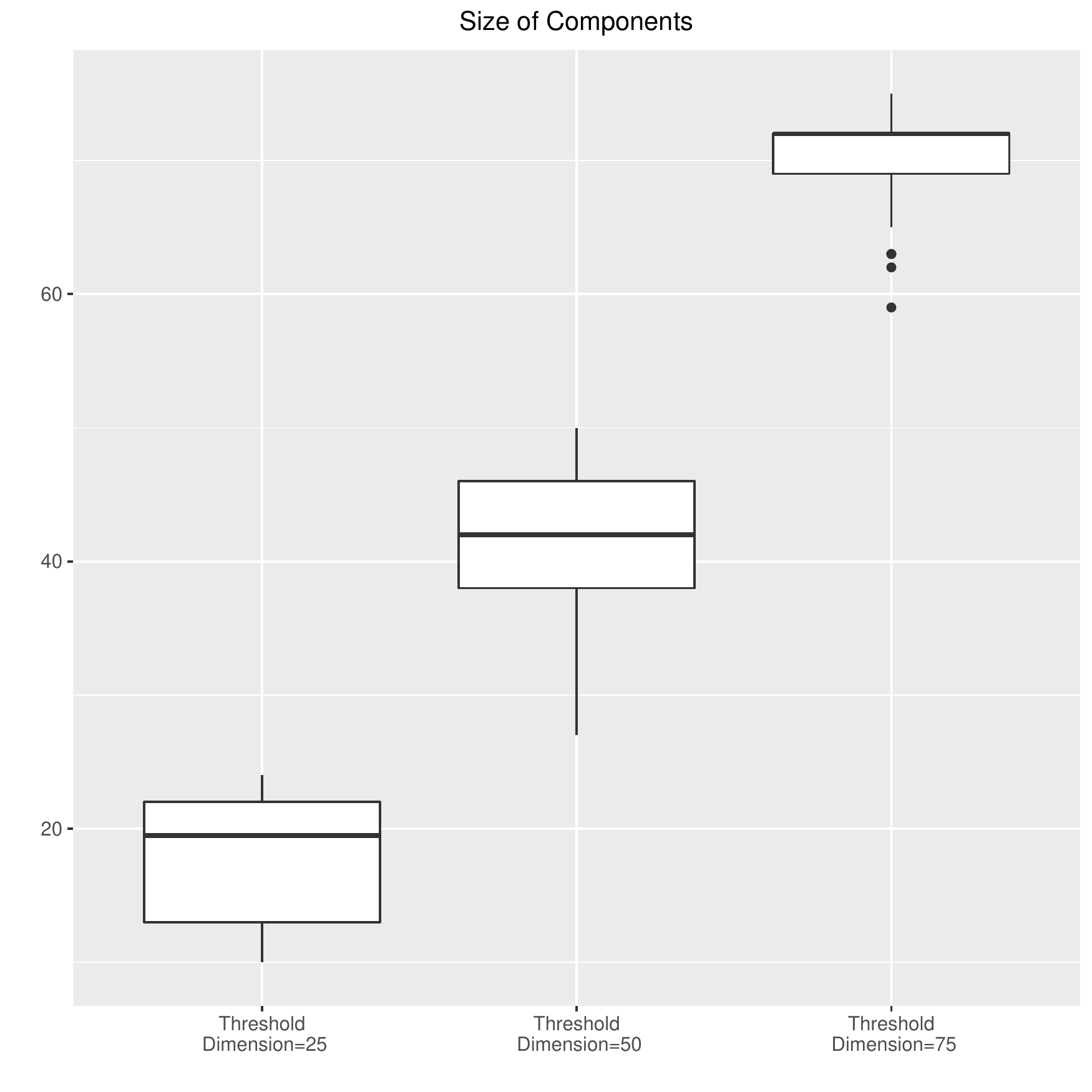}
	\includegraphics[width=0.48\textwidth, trim={0.1cm 0.1cm 0.1cm 0.1cm},clip]{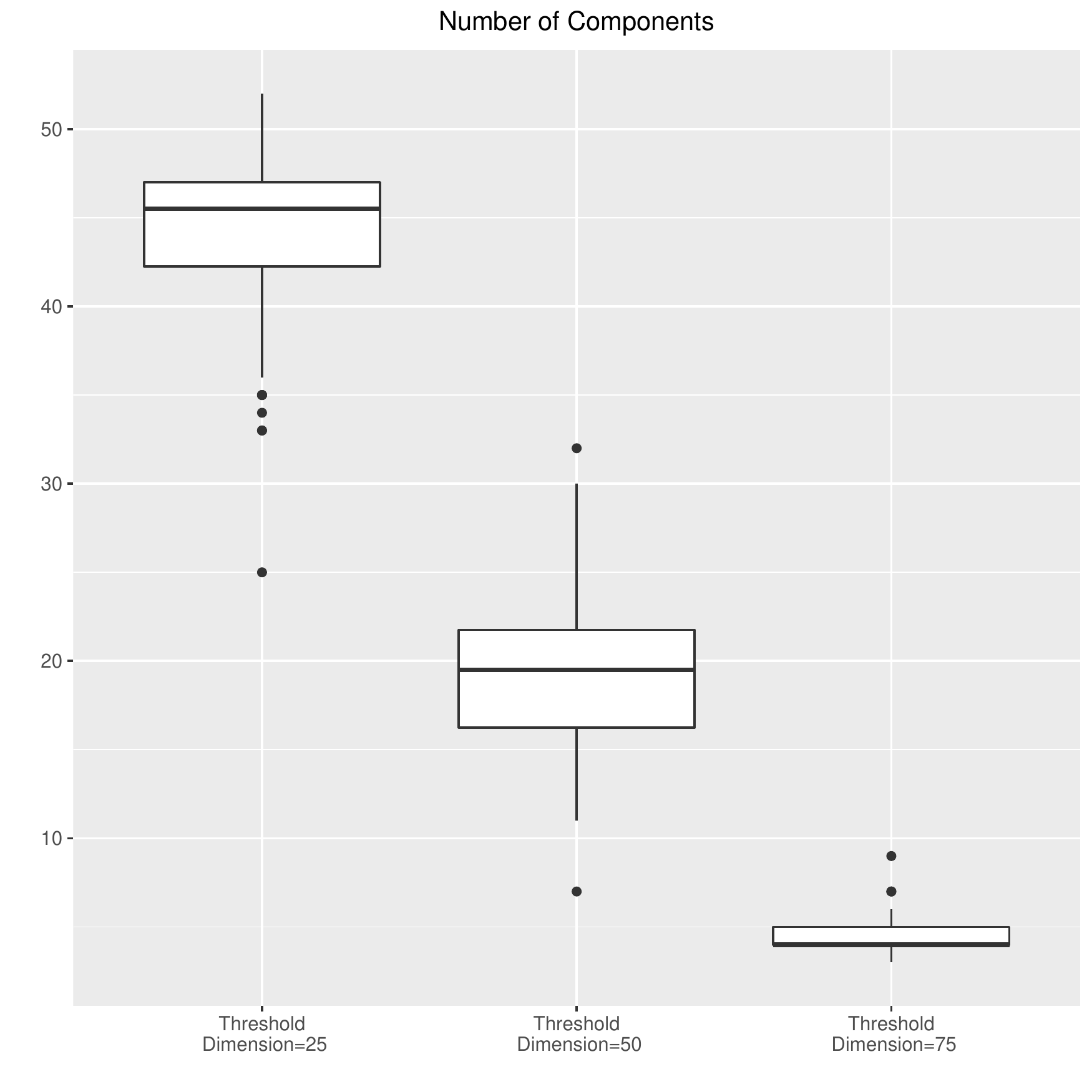}\\
	\caption{Scenario $V_1$, $2$-truncation: Distribution of maximal component sizes $\delta_T^i$ (left) and distribution of the number of connected components $p_T^i$ (right) for each of the $50$ replications, $i=1,\dots,50$ for different threshold dimensions $d_T=25,50,75$.}
	\label{fig:simstudy:clusterinfo}
\end{figure}

\begin{figure}[h]
	\centering
	\includegraphics[width=0.48\textwidth, trim={0.1cm 0.1cm 0.1cm 0.1cm},clip]{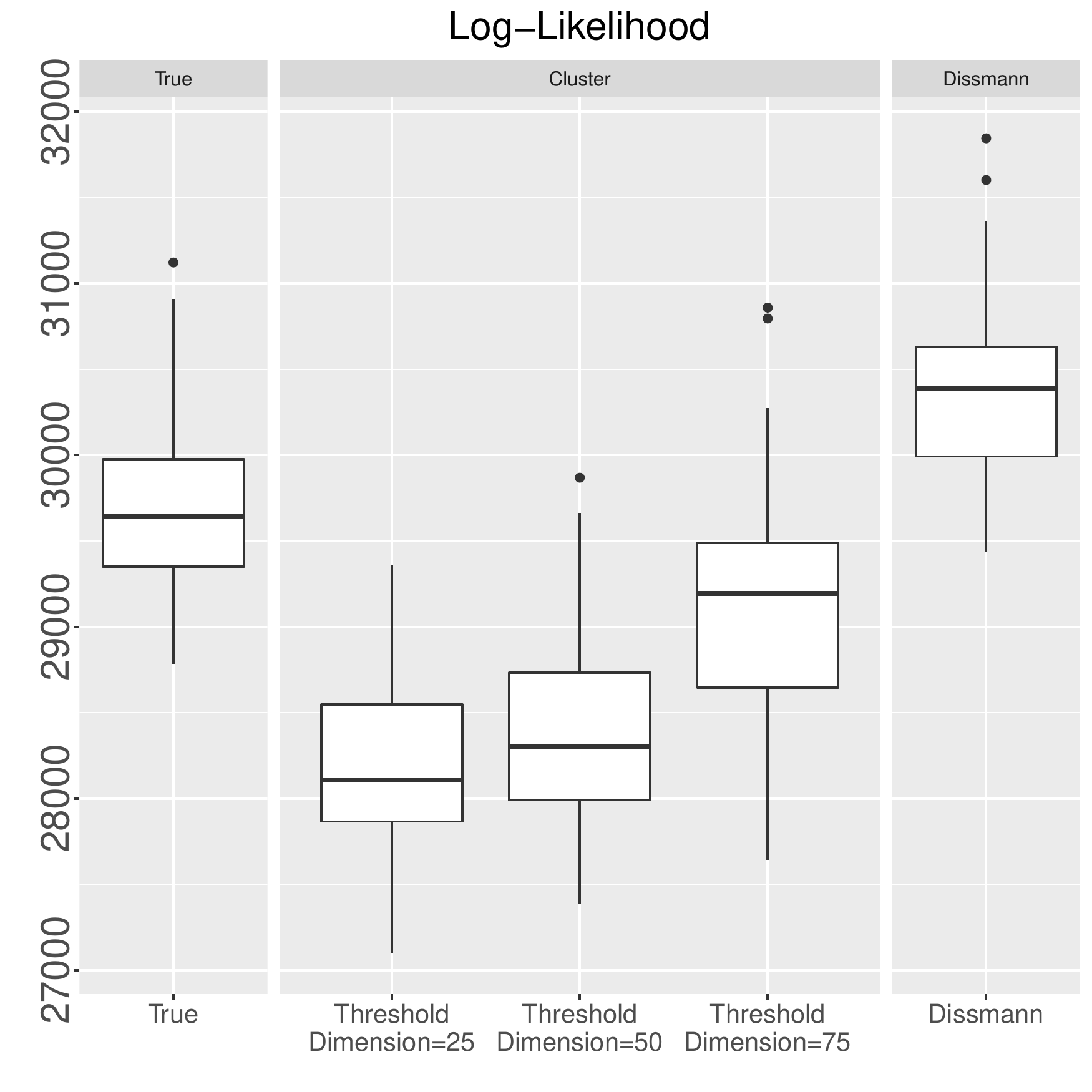}
	\includegraphics[width=0.48\textwidth, trim={0.1cm 0.1cm 0.1cm 0.1cm},clip]{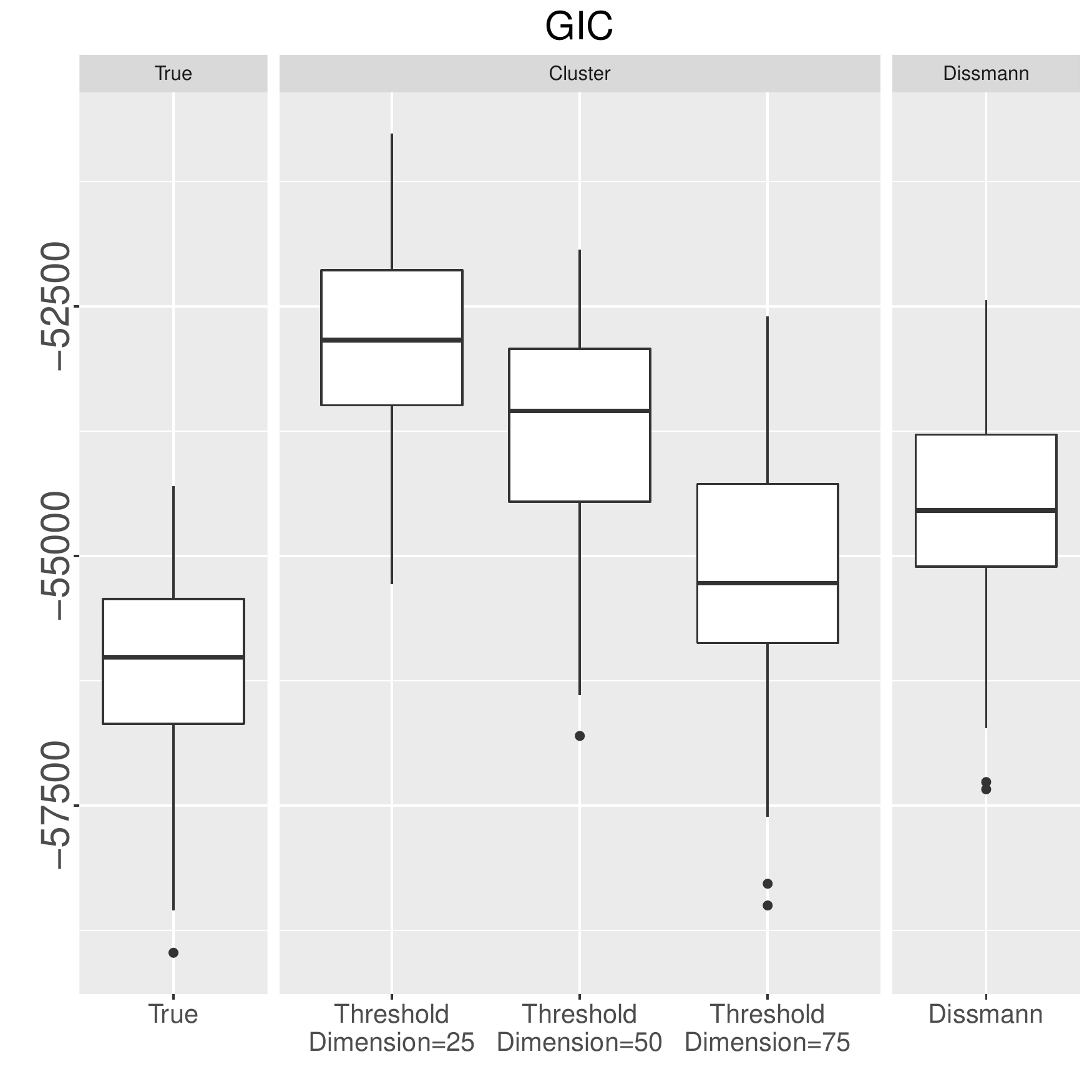}\\
	\includegraphics[width=0.48\textwidth, trim={0.1cm 0.1cm 0.1cm 0.1cm},clip]{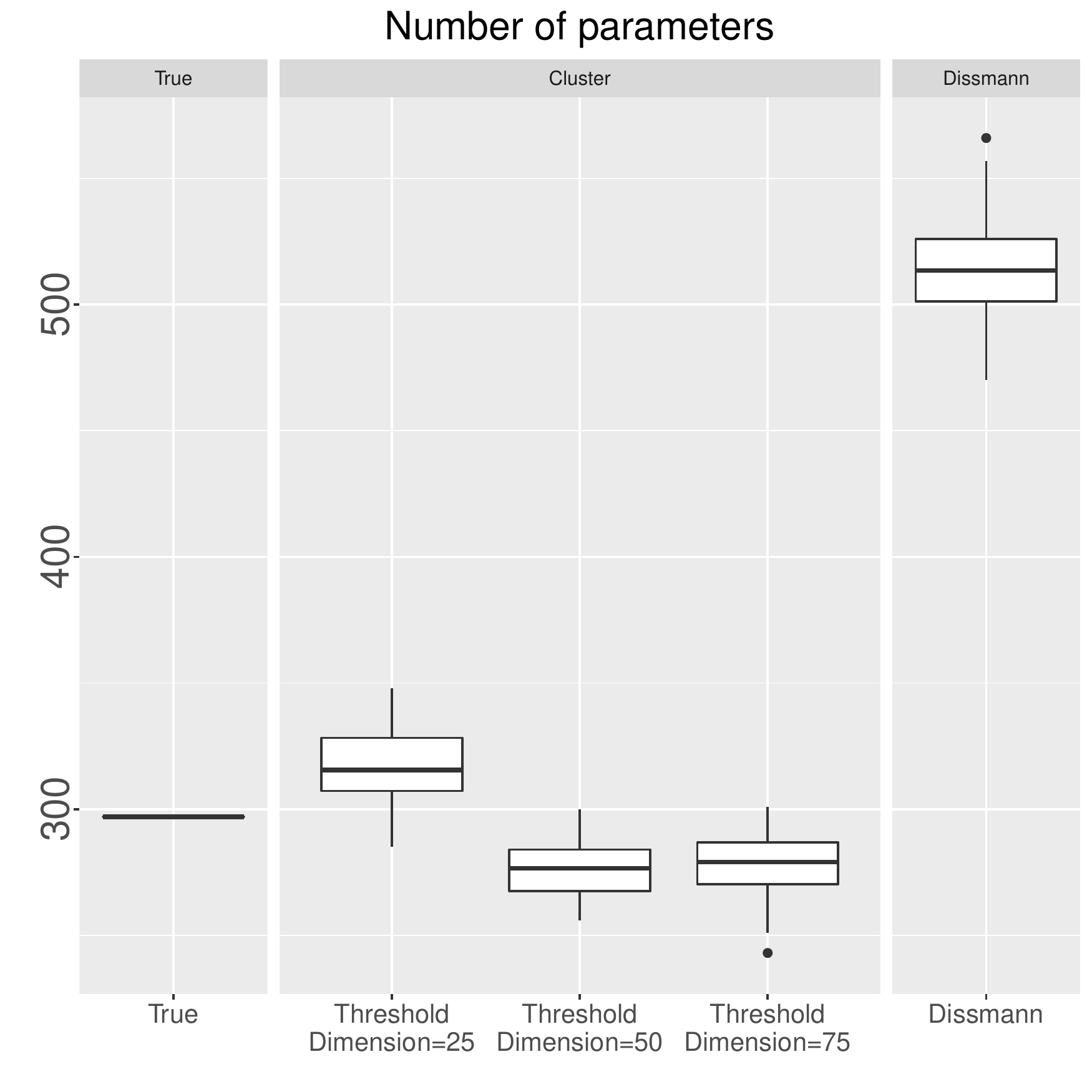}
	\includegraphics[width=0.48\textwidth, trim={0.1cm 0.1cm 0.1cm 0.1cm},clip]{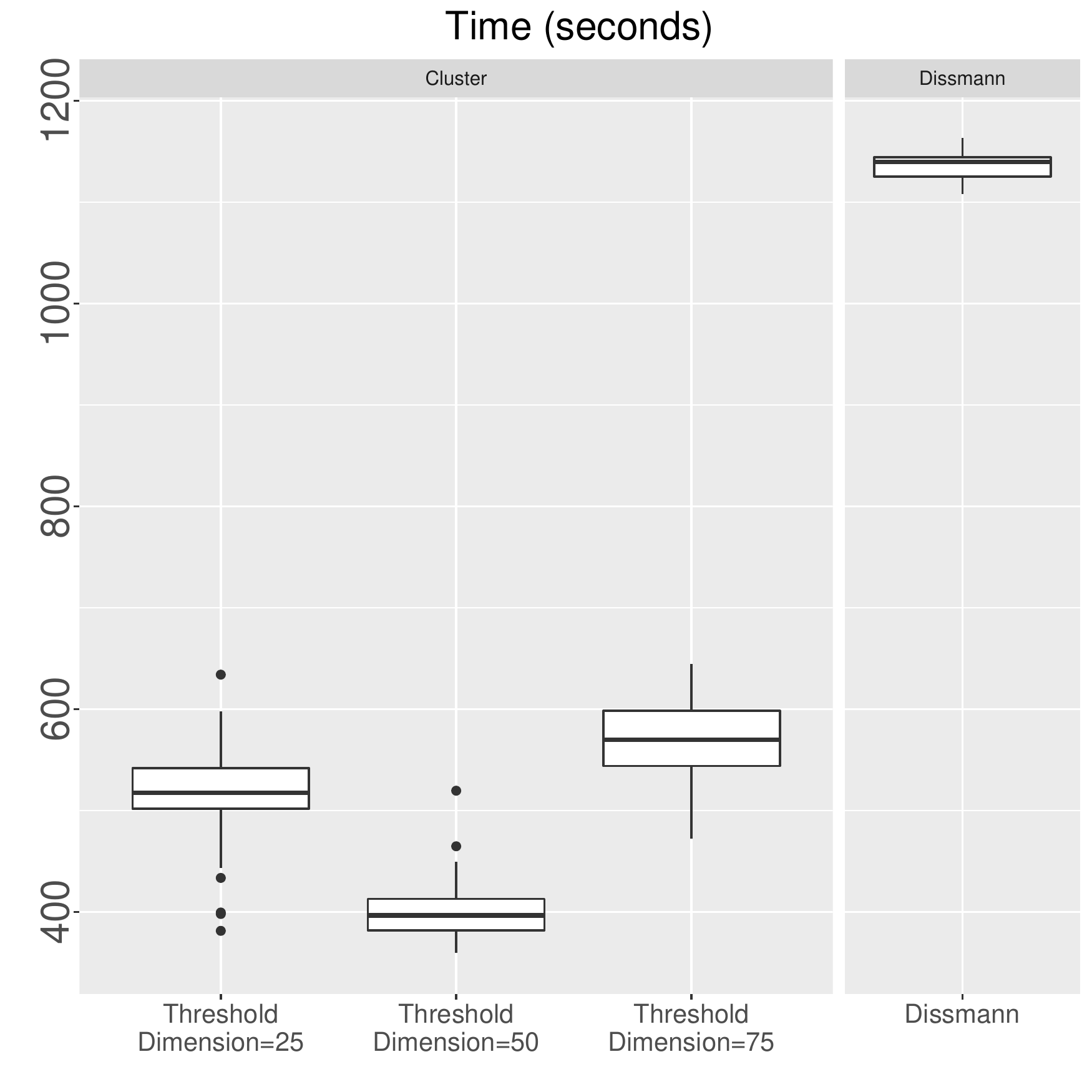}
	\caption{Scenario $V_1$, $2$-truncation: Comparison of \texttt{RVineClusterSelect} algorithm with threshold dimension $d_T = 25,50,75$ and Dissmann's algorithm on \textit{u-scale}: log-likelihood (upper left), GIC (upper right), number of parameters (lower left) and computation time (lower right).}
	\label{fig:simstudy:results2}
\end{figure}

First of all, we want to see whether our approach generates feasible models in moderate dimensions, e.\,g.\ $d \sim 100$. There, also the standard algorithm for vine copula estimation of \citet{dissmann-etal} works well. However, as our approach is especially targeted to sparse scenarios, we want to account for this and design three scenarios. We obtained stock quotes for $d=85$ stocks in the S\&P100 index from 01.01.2013 to 31.12.2016 and process them as previously described. The remaining $15$ stocks are removed since they dropped out of the index by the end of the observation period. We fit three different vine copula models $V_1,V_2,V_3$ to the data allowing for all parametric pair copula families implemented in the \texttt{VineCopula} R-package of \citet{vinecopula}, however imposing a $2$, $5$ and $10$-truncation. Additionally, we perform an independence test at significance level $\alpha = 0.05$ to pair copulas to the independence copula. Clearly, the $2$-truncation is the most parsimonious model, however, also the $5$ and $10$-truncation are quite reduced models compared to a full model with $84$ trees. From these fitted models, we simulate $50$ replications with $n=1000$ observations each for which we compare our method with the algorithm of Di{\ss}mann. We use runs with threshold dimensions $d_T = 25,50,75$, see \eqref{eq:thresholddim} and $k_F = \lceil\log\left(d\right)\rceil = \lceil\log\left(85\right)\rceil = 5$. This choice worked very well in the numerical experiments we carried out. Before evaluating the corresponding models, we consider the maximal component sizes $\delta_T^i$ and the number of connected components $p_T^i$ of the chosen partitions $T_i$ for all $50$ replications $i=1,\dots,50$. There, we observe that for higher threshold dimension, the number of the connected components decrease while the maximal component sizes naturally increases, see Figure \ref{fig:simstudy:clusterinfo}.
Next, we present the results with respect to goodness of fit, number of parameters and computation time for the $2$-truncation in Figure \ref{fig:simstudy:results2} and defer the remaining scenarios to the Appendix, Section \ref{sec:appendix:sim} as the results behave quite similarly.  We denote the model from which we simulate as true model.
First of all, we see that Di{\ss}mann's algorithm obtains highest log-likelihood in all scenarios. However, it also tends to overfit as it has almost twice as much parameters as the true model. Our approach captures the number of parameters better and this is accompanied with lower, i.\,e.\ better GIC, see \eqref{eq:AICBICGIC}, for the threshold dimension $d_T = 75$. We also see significantly lower computation times. Generally, our approach needs far less parameters but selects the most important bivariate dependencies to yield parsimonious models. 

\subsection{Runtime and Goodness of Fit Analysis on Real World Data}
\begin{figure}[ht]
	\centering
	\includegraphics[width=0.64\textwidth, trim={0.1cm 0.1cm 0.1cm 0.1cm},clip]{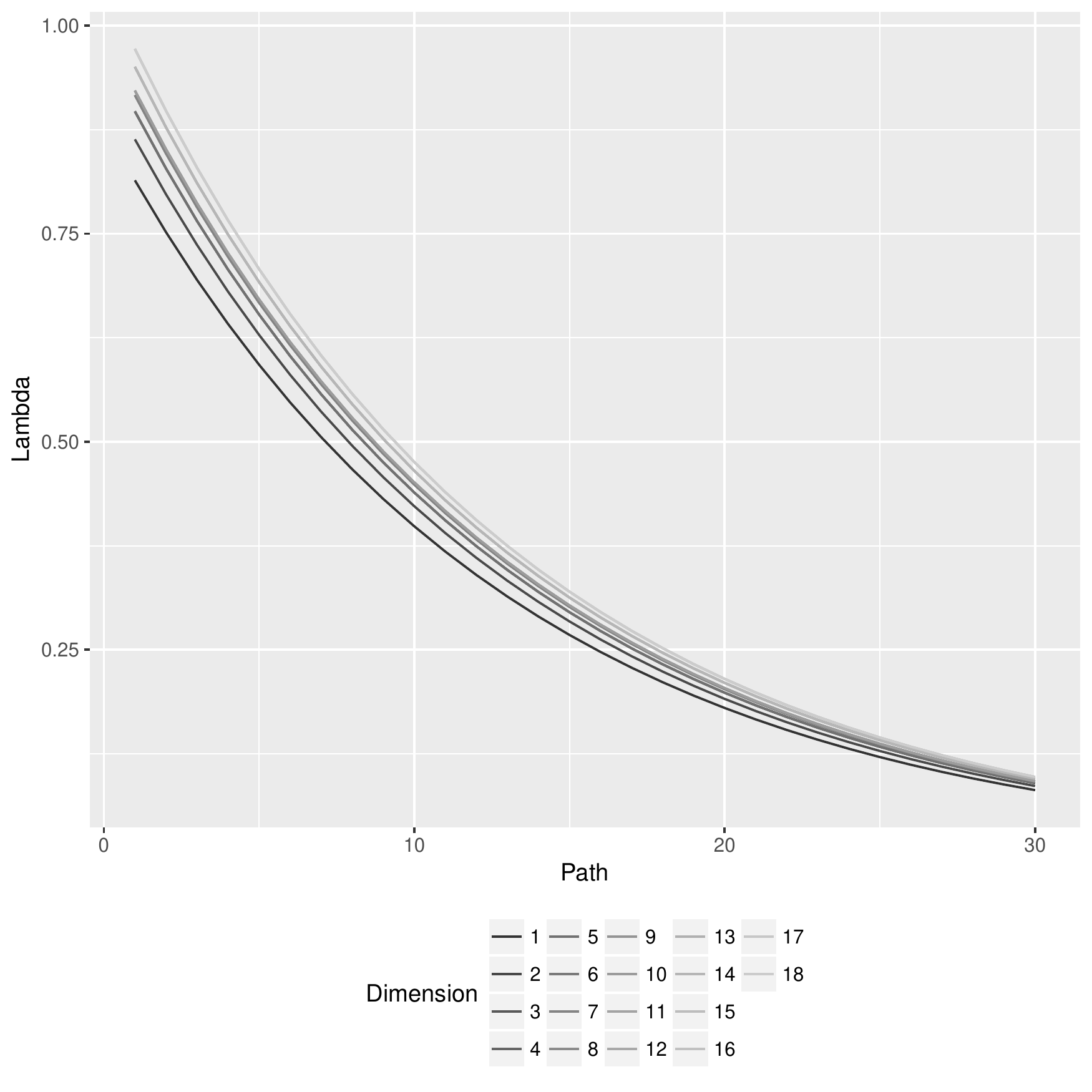}
	\caption{Comparison of $\bm{\lambda}_1,\dots,\bm{\lambda}_{18}$ for $U_1,\dots,U_{18}$ as calculated by the \texttt{huge} package for path length $J=30$.}
	\label{fig:lambda}
\end{figure}

\begin{figure}[H]
	\centering
	\includegraphics[width=0.49\textwidth, trim={0.1cm 0.1cm 0.1cm 0.1cm},clip]{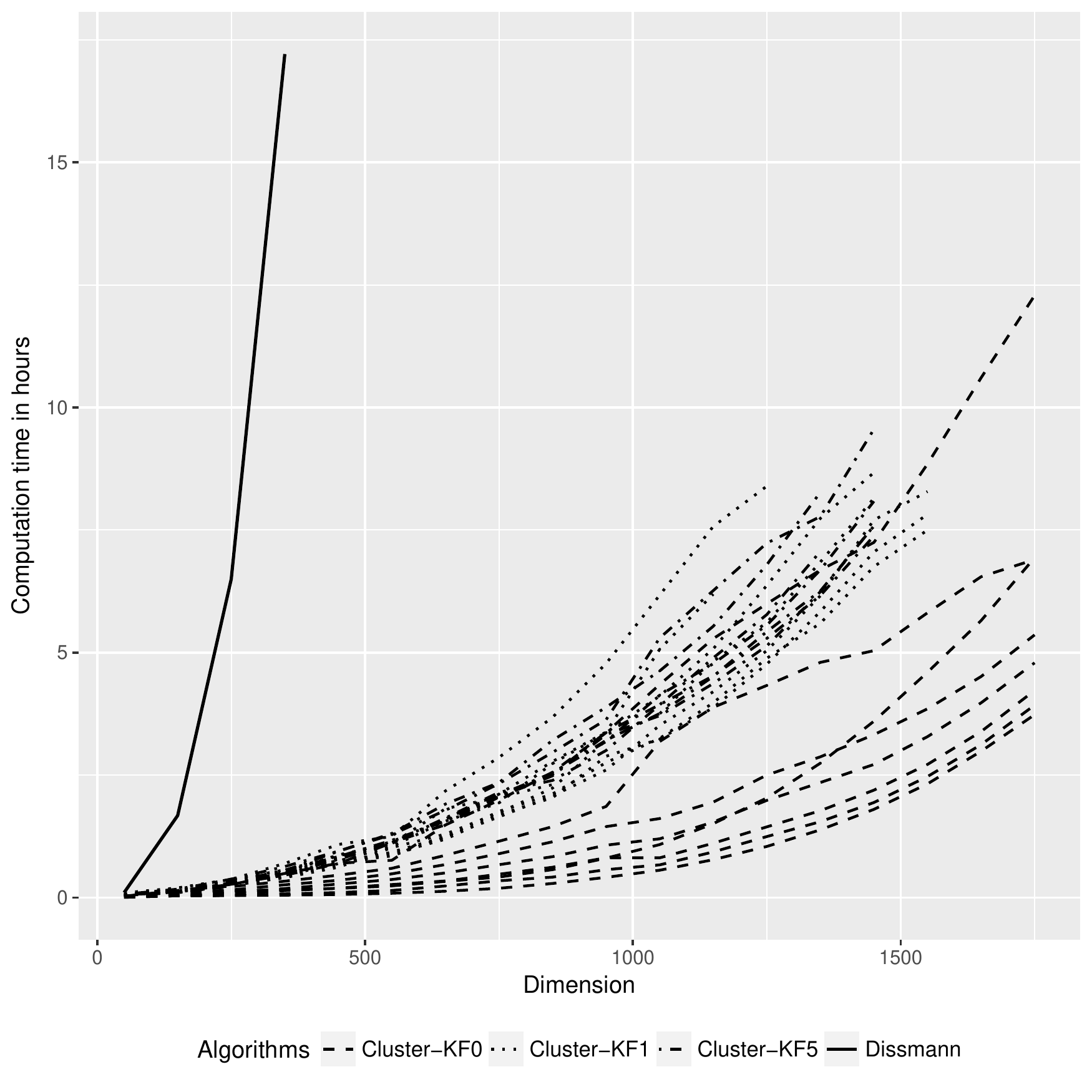}
	\includegraphics[width=0.49\textwidth, trim={0.1cm 0.1cm 0.1cm 0.1cm},clip]{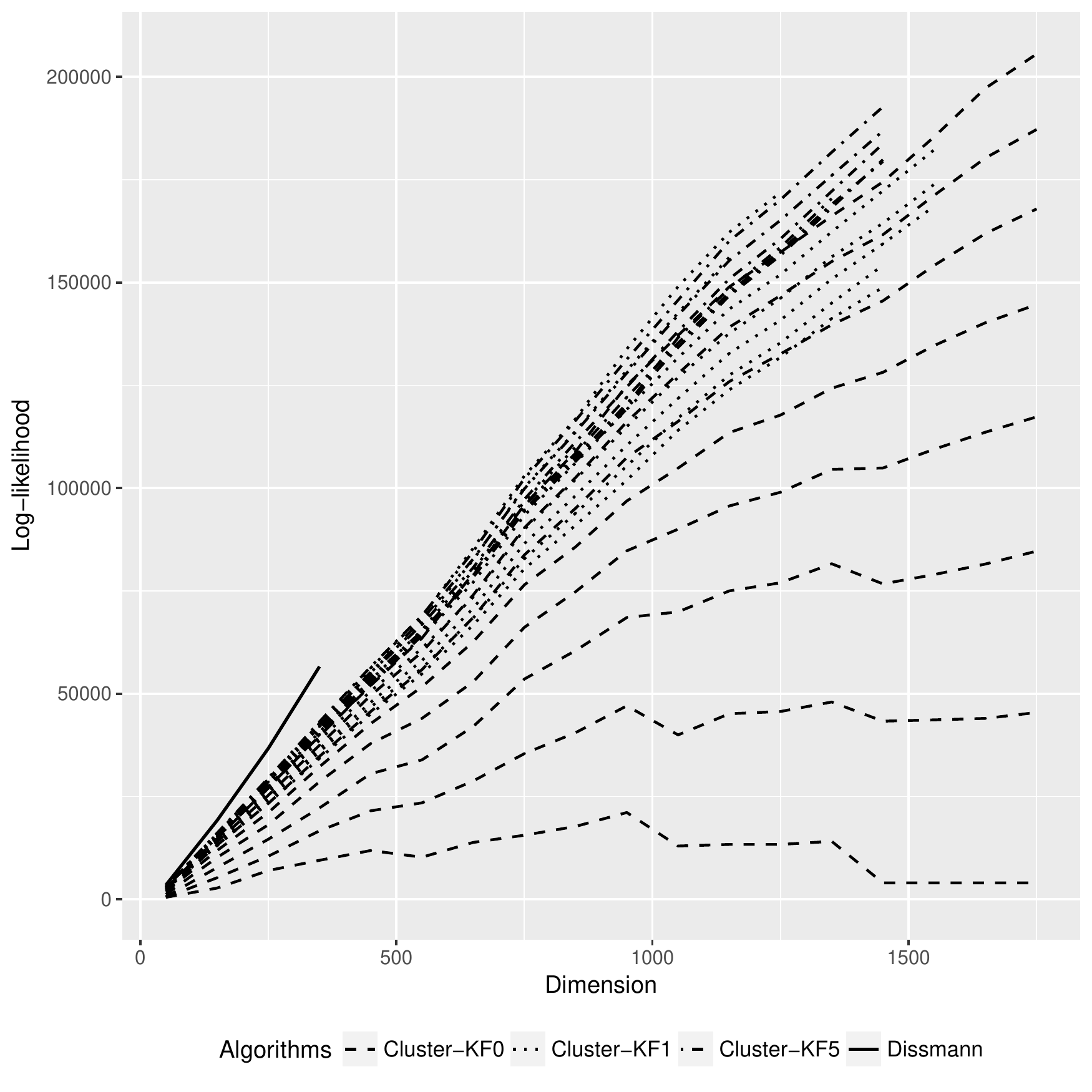}\\
	\includegraphics[width=0.49\textwidth, trim={0.1cm 0.1cm 0.1cm 0.1cm},clip]{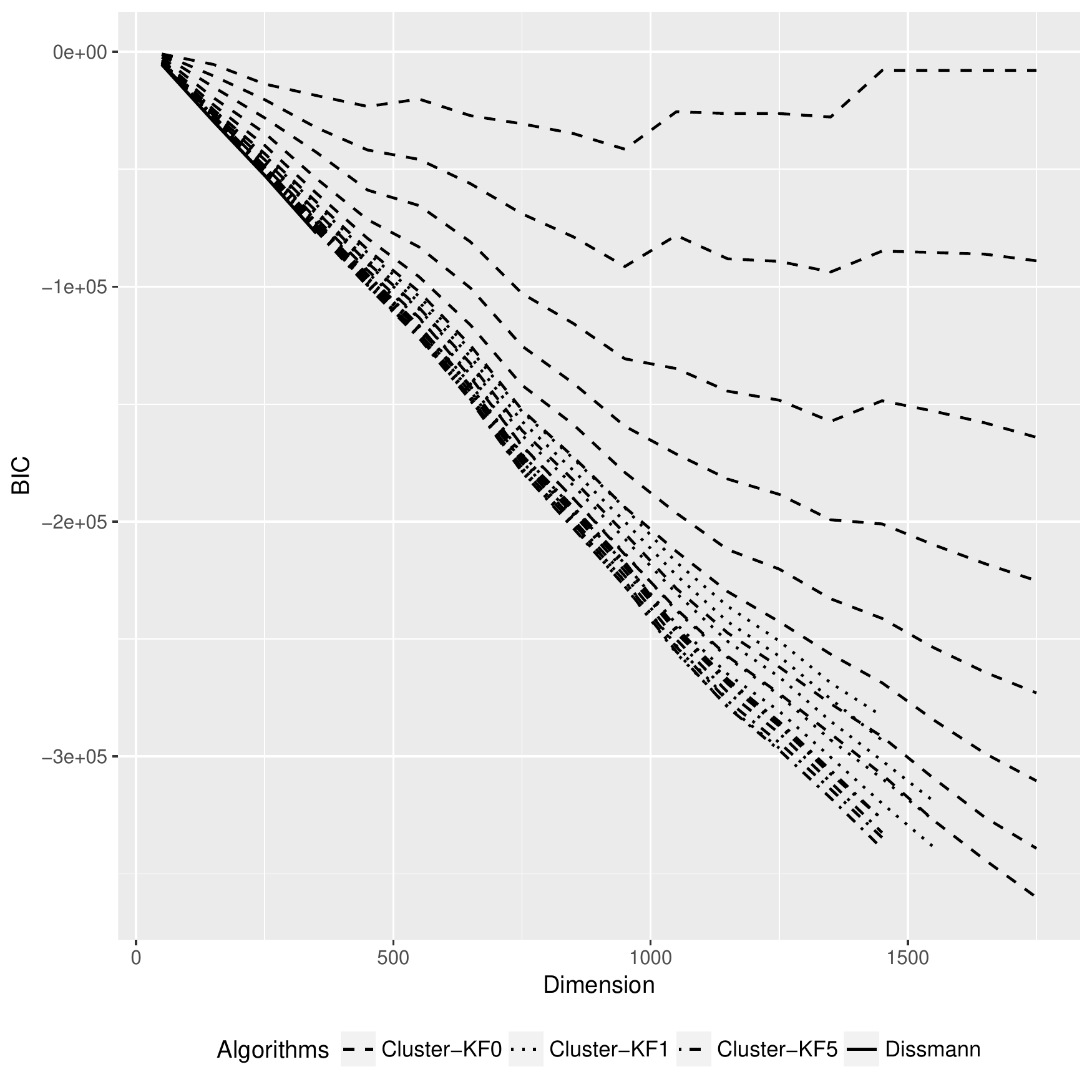}
	\includegraphics[width=0.49\textwidth, trim={0.1cm 0.1cm 0.1cm 0.1cm},clip]{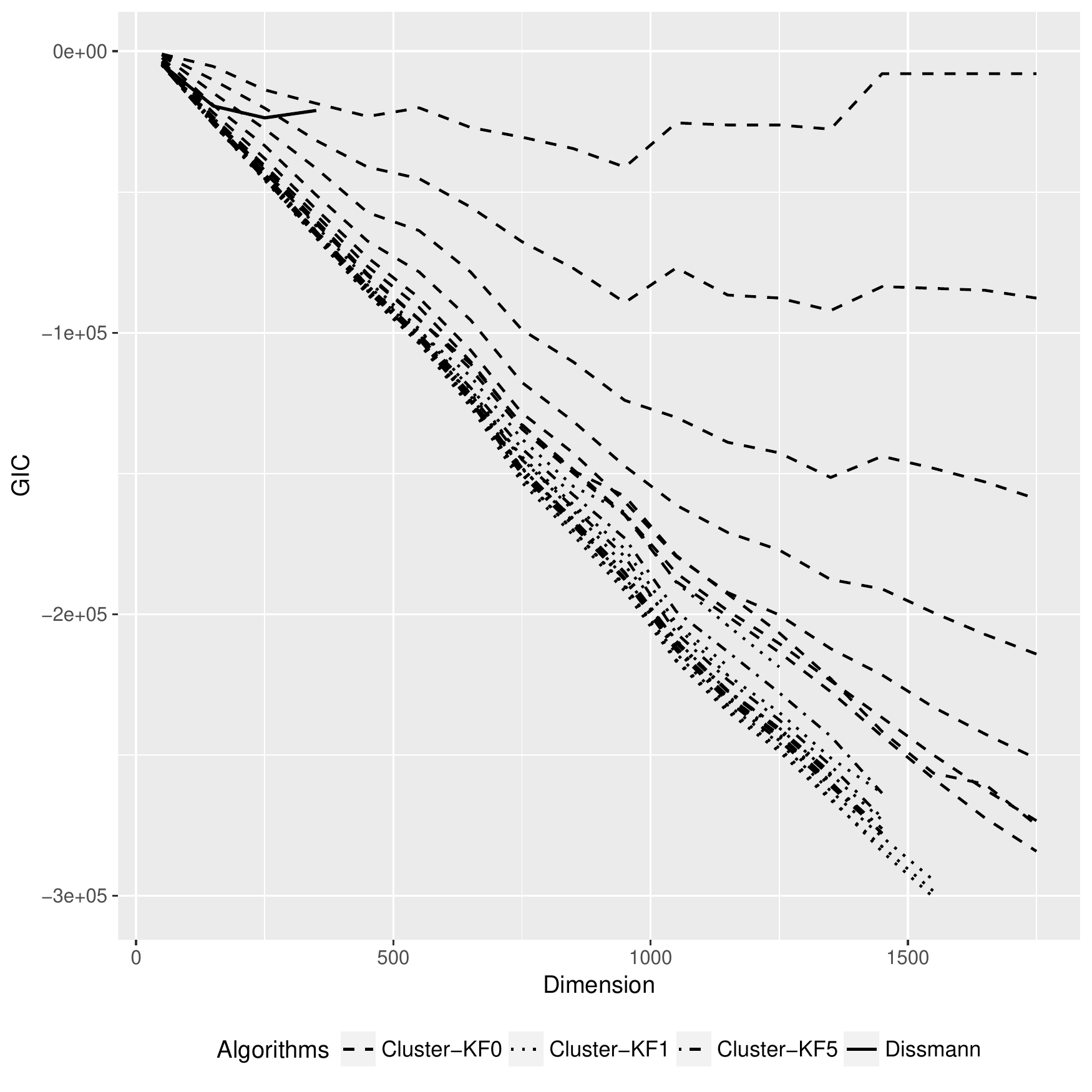}
	\caption{Comparison of \texttt{RVineClusterSelect} algorithm for $\GGG^\ell_2,\dots,\GGG^\ell_9$ with $k_F=0,1,5$ and Dissmann's algorithm on $d$ dimensions, $d=50,150,\dots,1750$: computation time (upper left), log-likelihood (upper right), BIC (lower left), GIC (lower right).}
	\label{fig:hdcomparison1}
\end{figure}

We obtained data from $d=1757$ stock listed companies from the entire globe in the financial services industry. The data contains $n = 470$ trading days, i.\,e.\ about two years of data. Thus, we have a data matrix $U \in \left[0,1\right]^{n \times d}$ and split it into $18$ nested subsets $U_\ell \in \left[0,1\right]^{n \times d_\ell}$ with $d_\ell \in \left\{50,150,250,\dots,1750\right\}$ by only considering the first $d_\ell$ columns. Our goal is to compare our approach to Di{\ss}mann's method considering the computation times with respect to the dimension. We expect that our approach is on the one hand side much faster, and on the other hand side more precise because of using a more accurate estimation approach as introduced in Section \ref{subsec:improvingaccuracy}. The question arises, which maximal component size $d_\ell^T$ to choose for our algorithm and which fill level $k_F$. For the maximal component size, we evaluate a grid of $9$ graphs with the maximal component sizes given Table \ref{tab:compsize}. Here, we refer to the maximal component size of graph $\GGG_j^\ell$ estimated on the data matrix $U_\ell$ by $\delta_j^\ell$, $\ell=1,\dots,18$ and $j=2,\dots,9$. Note that we do not select a threshold dimension upfront but take it as generated by the estimation process of the \texttt{huge} package.
\begin{table}[h]
	\centering
	\begin{tabular}{cc|rrrrrrrr}
		\hline
		Data matrix & Dimension $d_\ell$ & $\delta_2^\ell$ & $\delta_3^\ell$ & $\delta_4^\ell$ & $\delta_5^\ell$ & $\delta_6^\ell$ & $\delta_7^\ell$ & $\delta_8^\ell$ & $\delta_9^\ell$ \\ 
		\hline
		$U_1$ & 50 & 3 & 3 & 4 & 8 & 10 & 15 & 18 & 22 \\ 
		$U_2$ & 		150 &9 & 17 & 20 & 36 & 46 & 53 & 57 & 63 \\ 
		$U_3$ & 		250 &18 & 29 & 36 & 49 & 61 & 75 & 84 & 100 \\ 
		$U_4$ & 		350 &26 & 40 & 52 & 71 & 90 & 105 & 121 & 172 \\ 
		$U_5$ & 		450 &24 & 34 & 64 & 86 & 107 & 134 & 152 & 205 \\ 
		$U_6$ & 		550 &17 & 38 & 67 & 102 & 127 & 148 & 175 & 203 \\ 
		$U_7$ & 		650 &20 & 43 & 84 & 111 & 154 & 181 & 210 & 307 \\ 
		$U_8$ & 		750 &20 & 43 & 105 & 144 & 187 & 212 & 248 & 359 \\ 
		$U_9$ & 		850 &21 & 55 & 75 & 160 & 197 & 239 & 278 & 406 \\ 
		$U_{10}$ & 		950 &22 & 59 & 82 & 176 & 220 & 265 & 307 & 452 \\ 
		$U_{11}$ & 		1050 &12 & 50 & 88 & 106 & 224 & 254 & 435 & 494 \\ 
		$U_{12}$ & 		1150 &12 & 54 & 93 & 112 & 238 & 270 & 458 & 530 \\ 
		$U_{13}$ & 		1250 &12 & 54 & 93 & 112 & 238 & 270 & 470 & 545 \\ 
		$U_{14}$ & 		1350 &12 & 54 & 92 & 113 & 239 & 272 & 483 & 562 \\ 
		$U_{15}$ & 		1450 &3 & 49 & 81 & 108 & 224 & 267 & 465 & 570 \\ 
		$U_{16}$ & 		1550 &3 & 49 & 81 & 108 & 228 & 271 & 477 & 618 \\ 
		$U_{17}$ & 		1650 &3 & 49 & 81 & 108 & 228 & 272 & 490 & 654 \\ 
		$U_{18}$ & 		1750 &3 & 49 & 81 & 108 & 228 & 272 & 439 & 680 \\ 
		\hline
	\end{tabular}
\caption{Maximal component sizes $\delta^\ell_2,\dots,\delta^\ell_9$ of graphs $\GGG^\ell_2,\dots,\GGG^\ell_9$, for data matrices $U_\ell$, see \eqref{eq:graphseq} for each dimension $d_\ell$, $\ell=1,\dots,18$.}
\label{tab:compsize}
\end{table}
The corresponding values of $\lambda$ calculated for each of the $18$ different data matrices behave quite similar, see Figure \ref{fig:lambda}.
We omit the first graph $\GGG_1$ since it is always an empty graph, and hence, there are no connected components and only isolated nodes. For the fit based on the associated graphs, we evaluate the fill levels $k_F = 0,1,5$. Note however that we have to be cautious here. If $k_F=1$, by our algorithm, for the first tree, the worst case effort can only be bounded by $d^2$ as no independence information is used, which may take some time. It is hence more computationally feasible to only fit for example one-parametric pair copula families \textit{outside} the connected components to ease computational effort.
Our findings are given in Figure \ref{fig:hdcomparison1}, describing the connection between dimension of the sample and computation time, log-likelihood, BIC and GIC. Here, we denote our algorithm by \texttt{Cluster-KF} with the corresponding fill-levels $k_F = 0,1,5$, dashed, dotted and dot-dashed, respectively. Note that Di{\ss}mann's algorithm in the BIC plot is below all other plots.
In total, we have one line for Di{\ss}mann's algorithm and eight lines for our approach with different maximal cluster sizes as displayed in Table \ref{tab:compsize}. We perform all the calculations on a Linux Cluster with $32$ cores. We see that for computation time, Di{\ss}mann's algorithm computation time grows very rapidly once dimension exceeds $d > 100$. There are only $4$ data points since for the fifth estimation with $d=450$, we could not fit any model since we got a time out after $6$ days. Additionally, memory consumption becomes also a burden in these dimensions. For our proposed approach, the slope is much less step and fitting a model in the full size of $d=1750$ only takes about $4$ to $5$ hours on the Linux cluster with $k_F=0$. We note that a full maximum likelihood fitting for each pair copula was carried out and not just a mere estimation by inversion of the empirical Kendall's, which is faster. We report however, that also Di{\ss}mann's algorithm with this faster estimation is not feasible once the dimension exceeds about $d = 400$. Even though log-likelihood is larger for the standard algorithm, we attain equally good results in terms of BIC and even better results in terms of GIC. This is especially beneficial since we expect the overall dependence to decrease as dimension increases. This is because we expect that most intra-sectoral and intra-geographical dependencies are modelled first, and afterwards, we have more conditional independence. Thus, the number of parameters is expected not to grow as much as the dimension, expressed by lower GIC which penalizes complex models more than BIC. For $k_F=1$ and $k_F=5$, some data points in high dimensions are missing because of out-of-memory problems, which is related to many pair copula fits for higher fill levels in dimensions $d > 1500$. However, we see that even though computation time is increasing, it is acceptable given the large benefits in terms of log-likelihood and GIC.

\subsection{Data Application}
\begin{figure}[h]
	\centering
	\includegraphics[width=0.49\textwidth, trim={0.1cm 0.1cm 0.1cm 0.1cm},clip]{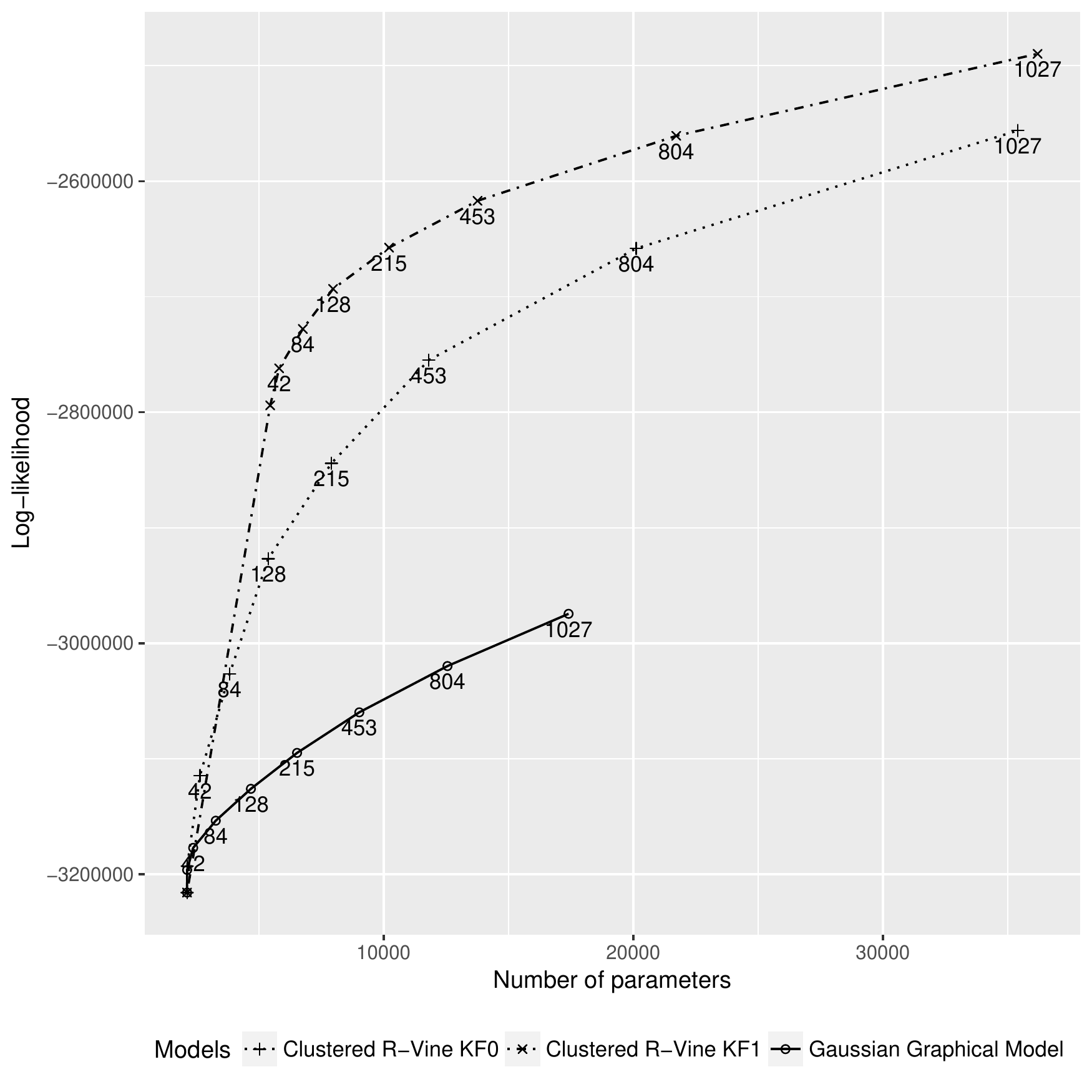}
	\includegraphics[width=0.49\textwidth, trim={0.1cm 0.1cm 0.1cm 0.1cm},clip]{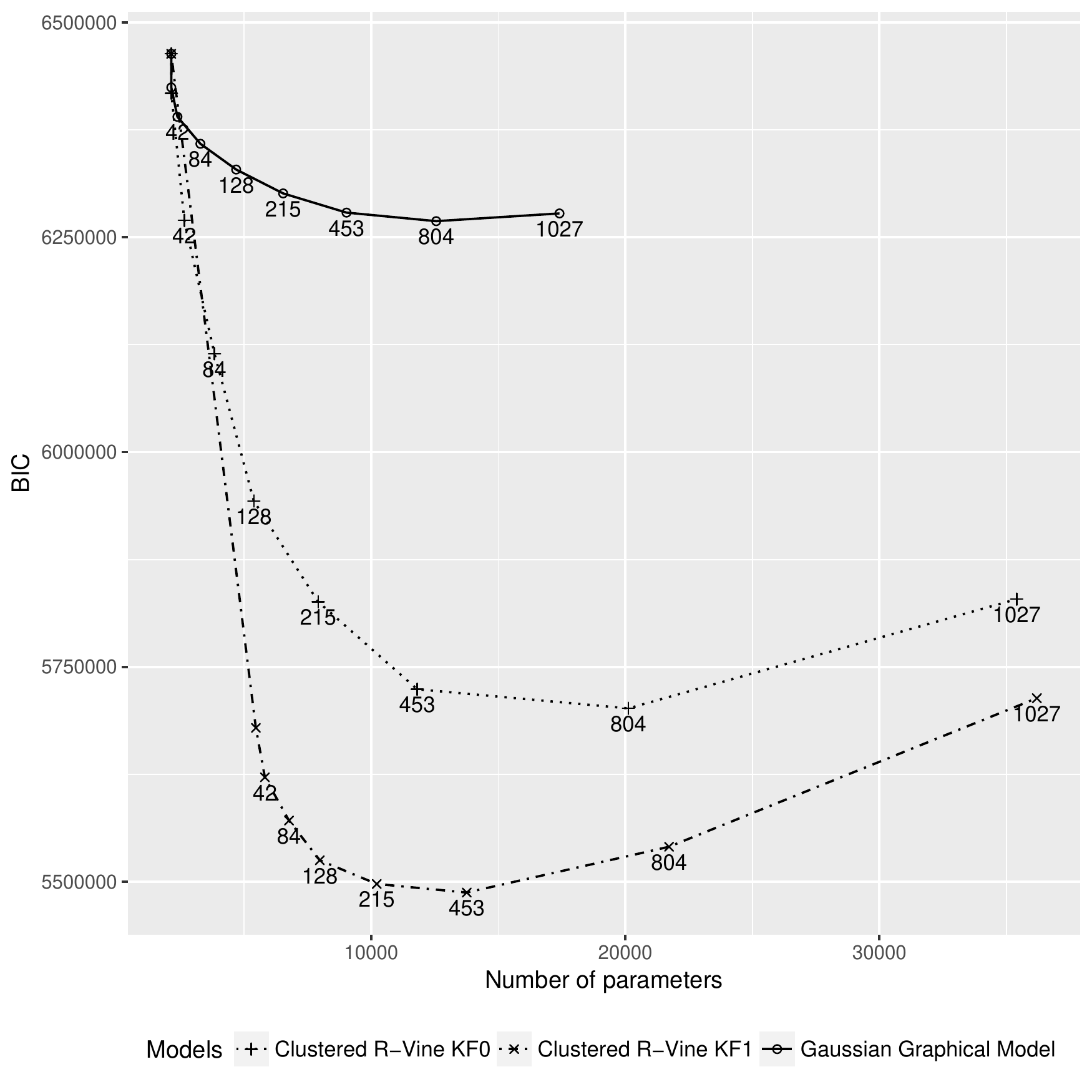}
	\caption{Comparison of clustered R-vines with fill-levels $k_F=0,1$ and Gaussian Graphical Model fitted with the Graphical Lasso $\GGG_1,\dots,\GGG_9$: log-likelihood (left), GIC (lower right).}
	\label{fig:datacomparison}
\end{figure}

\begin{figure}[h]
	\centering
	\includegraphics[width=0.48\textwidth, trim={0.1cm 0.1cm 0.1cm 0.1cm},clip]{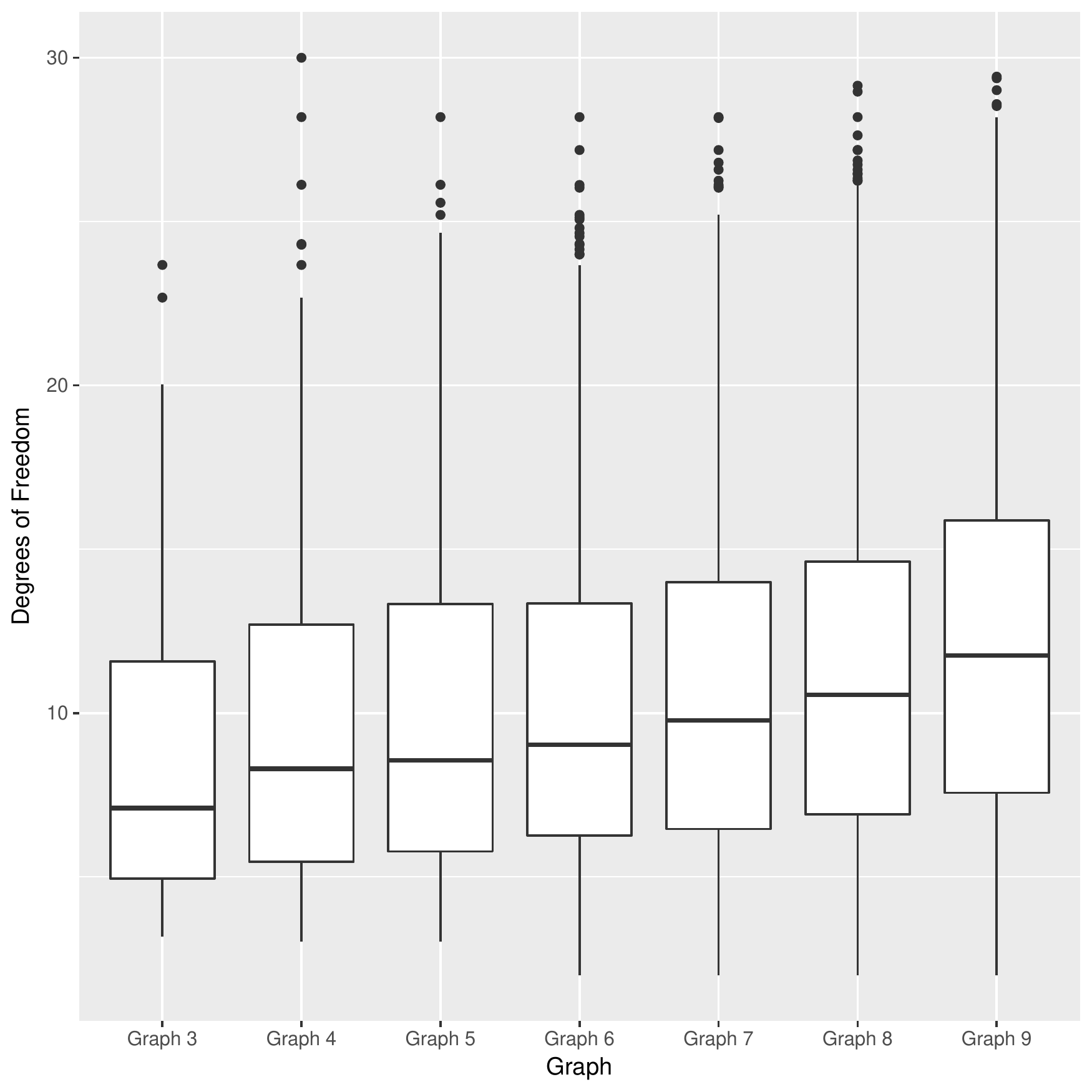}
	\caption{Distribution of the degrees of freedom as parameter for the Student's-t copula.}
	\label{fig:tcopuladf}
\end{figure}

If we want to compare the Gaussian modeling approach with the non-Gaussian R-vine approach, we need to set the fill-level $k_F = 0$, i.\,e.\ only model the edges present in the Gaussian model by pair copulas and not include additional edges between the connected components. However, we will also include a fit using fill-level $k_F=1$ in our considerations later on.
Secondly, we have to assume the same marginal distributions. This time, we obtained data on $d=2131$ stocks with $n=999$ observations. The stocks are based in the US (1866), Sweden (154) and Australia (111). The following industry sectors are covered, see Table \ref{table:overviewsectors}.
\begin{table}[h]
	\centering
	\begin{tabular}{rlllll}
		\hline
ID & Sector Description & Total & USA & Sweden & Australia \\ 
\hline
 1 & Materials and Energy & 218 & 209 & 3 & 6 \\ 
 2 & Industry Conglomerates & 4 & 4 & 0 & 0 \\ 
 3 & Consumer Staples & 205 & 189 & 12 & 4 \\ 
 4 & Financial Services and Real Estate & 524 & 449 & 49 & 26 \\ 
 5 & Healthcare and Chemicals & 232 & 182 & 26 & 24 \\ 
 6 & Manufacturing, Industrials and Defense & 195 & 173 & 12 & 10 \\ 
 7 & Business Management and Services & 325 & 295 & 21 & 9 \\ 
 8 & IT, Telecommunication and Software & 360 & 301 & 28 & 31 \\ 
 9 & Utilities & 68 & 64 & 3 & 1 \\ 
		\hline \hline
		& & 2131 & 1866 & 154 & 111
	\end{tabular}
\caption{Distribution of stocks over industry sectors and geographies in data application.}
\label{table:overviewsectors}
\end{table}
Define $S_i$ the $i$-th stock for $i=1,\dots,2131$, denote the industry sector assignment by
\begin{equation*}
\III_\ell: \left\{i: \mbox{Stock } S_i \mbox{ belongs to industry } \ell \right\} \mbox{ for } \ell = 1,\dots,9,
\end{equation*}
and the node set $V = \left\{1,\dots,2131\right\}$ for all graphical models $\GGG_1,\dots,\GGG_{30}$. We estimate a path of $30$ graphical Lasso solutions $\GGG_1,\dots,\GGG_{30}$ calculated by the \texttt{huge} R-package without setting a threshold dimension. To obtain also the log-likelihoods, we evaluate the corresponding covariance matrices of the solutions. We report the first ten corresponding maximum connected component sizes of $\GGG_1,\dots,\GGG_{10}$ in Table \ref{table:maxconnectedcompsize}.
\begin{table}[h]
	\centering
	\begin{tabular}{c|c|c|c|c|c|c|c|c|c}
		$\delta_1$ & $\delta_2$ & $\delta_3$ & $\delta_4$ &	$\delta_5$ & $\delta_6$ & $\delta_7$ & $\delta_8$ & $\delta_9$ &  		$\delta_{10}$\\
		\hline
		1 & 2 & 42 & 84 & 128 & 215 & 453 & 804 & 1027 & 1231
	\end{tabular}
	\caption{Maximal connected component sizes for $\GGG_1,\dots,\GGG_{10}$.}
	\label{table:maxconnectedcompsize}
\end{table}
\begin{equation*}
\end{equation*}
The remaining 20 cluster sizes for $\GGG_{11},\dots,\GGG_{30}$ are significantly higher than $1200$ and thus not considered for our method, as these subsets are too large to fit an R-vine model onto them. In fact, we only fit R-vines on the graphs $\GGG_2,\dots,\GGG_9$ since the first partition with $\delta_1 = 1$ is an empty graph bearing no information and the largest $\GGG_{10}$ with $\delta_{10} = 1231$ is also too large. Hence, we consider $8$ graphical models $\GGG_2,\dots,\GGG_9$ in the following. An interesting property we can observe is the \textit{industry sector concentration} within each connected component of the graphical models $\GGG_2,\dots,\GGG_9$. In the notation of Section \ref{subsec:clustering}, consider graphical models $\GGG_j=\left(V^j = \bigcup_{i=1}^{p_j}~V_i^j,\bigcup_{i=1}^{p_j}~{E}_i^j\right)$ with for $j=2,\dots,9$. Let us fix $j$ and consider the $k$-th component $V_k^j$ of $V^j$. Then, denote $\left|V_k^j\right| = \nu_k^j$, i.\,e.\ $\nu_k^j$ stocks are contained in this connected component. Recall that each node represents a stock associated to one industry sector. We define the industry sector occurring most often in the component $V_k^j$ by
\begin{equation*}
b_k^j = \argmax_{\ell \in 1,\dots,9}~\sum_{i \in V_k^j}~\mathds{1}_{\left\{i \in \III_\ell\right\}}\left(i\right).
\end{equation*}
A natural measure for the sector concentration in $V_k^j$ is given by
\begin{equation}\label{eq:sectorconcentration}
\varrho_k^j = b_k^j / \nu_k^j.
\end{equation}
In other words, we count the occurrence of all different sectors within each connected component and divide the number of the most often occurring sector per connected component by the total number of nodes in this connected component. We do this for all connected components within each of the graphs in the sequence $\GGG_2,\dots,\GGG_9$. The results, see Table \ref{table:sectorconcentration}, demonstrate a very high sector concentration in the connected components over the entire sequence of considered graphs.
\begin{table}
\centering
\begin{tabular}{p{0.5cm}|p{0.5cm}|p{4.7cm}|p{4.7cm}|p{2.2cm}}
	\hline
	$\GGG_j$ & $p_j$ & $|\left\{k \in 1,\dots,p_j: \varrho_k^j = 1\right\}|$ & $|\left\{k \in 1,\dots,p_j: \varrho_k^j = 1\right\}|$ $/p_j$  & mean of $\left\{\varrho_k^j: \varrho_k^j \neq 1\right\}$ \\ 
	\hline
$\GGG_2$ & 3 & 3 & 1 & - \\ 
$\GGG_3$  & 20 & 19 & 0.95 & 0.83 \\ 
$\GGG_4$  & 42 & 39 & 0.93 & 0.64 \\ 
$\GGG_5$  & 67 & 62 & 0.93 & 0.74 \\ 
$\GGG_6$  & 69 & 60 & 0.87 & 0.66 \\ 
$\GGG_7$  & 61 & 46 & 0.75 & 0.69 \\ 
$\GGG_8$  & 47 & 41 & 0.87 & 0.55 \\ 
$\GGG_9$  & 31 & 26 & 0.84 & 0.58 \\ 
	\hline
\end{tabular}
\caption{Number of connected components $p_j$ in $\GGG_j$ for $j=2,\dots,9$, number of connected components $k=1,\dots,p_j$ with sector concentration $\varrho_k^j = 1$, percentage of connected components with sector concentration $\varrho^j_k=1$ compared to all connected components and mean of the sector concentrations $\varrho_k^j$ over the remaining connected components where sector concentration $\varrho_k^j \neq 1$.}
\label{table:sectorconcentration}
\end{table}
We see first of all, that a large portion of connected components have sector concentration $\varrho_j = 1$, second and third column of Table \ref{table:sectorconcentration}. Thus, at least $80 \%$ of the connected components are dominated by a single sector. Additionally, we see from the last column that also the connected components where more than one sector are present, have quite a high sector concentration. This backs our assumption that the graphical Lasso works very well to isolate highly dependent subsets from each other. The motivation for this is the assumption that especially in high dimensions, there exist \textit{clusters of dependence}. As we have e.\,g.\ geographical or industry-sectoral dependency in a high dimensional stocks dataset, we can expect the model to make use of \textit{conditional independences} or, in other words, \textit{sparsity}. The idea is that after all intra-geographical or sectoral dependencies are described, cross geographical or sectoral dependencies are weak and can be neglected outside the connected components. The same is often monitored for biological data, where only a small set of genes is affected by each other.\\
Finally, we compare the Gaussian model and the R-vine model in terms of log-likelihood and GIC on the \textit{z-scale} for corresponding numbers of parameters, see Figure \ref{fig:datacomparison}. In the Gaussian model, the number of parameters is equal to the numbers of edges in the graph whereas for the R-vine, the number of parameters is equal to the number of pair copula parameters. In both cases, we add $d$ parameters for the estimated variance of the marginal distributions. This is because we need to add Gaussian marginals with the same variance as estimated by the graphical Lasso estimate to our R-vine model to make it comparable to the Gaussian model estimated with the graphical Lasso.
We clearly see that the flexibility of the R-vine compared to the Gaussian model leads to significant out-performance with respect to log-likelihood and even more with respect to GIC. This stems from the fact that the R-vine gains much more exploratory power with adding additional parameters. This also true for the corresponding model with $k_F = 1$. The most parsimonious model in terms of GIC is the R-vine given by $\GGG_8$. This is also similar for the Gaussian model.\\
In terms of pair copulas, the most prominently present families in $\GGG_8$ are \textit{Frank-copulas} (5437, 39.9 \%) and \textit{Student's-t copulas} (1889, 13.8 \%) before \textit{Gaussian} (1581, 11.6 \%), which is a clear indicator of non-Gaussianity and also tail dependence. When considering the Student's-t copulas in the R-vine and their parameters, i.\,e.\ the degrees of freedom $df$, we have that for lower $df$, we have more tail dependence and for higher $df > 30$, the distribution becomes quite similar to the Gaussian distribution. We consider boxplots of the degrees of freedom over the R-vines computed with the graphs $\GGG_3,\dots,\GGG_9$, see Figure \ref{fig:tcopuladf}. Note that in the R-vine computed based on $\GGG_2$ we do not have any Student's-t copulas.
Since most of the degrees of freedom vary around $10$ to $15$, we clearly monitor and model heavy tailed data, which is quite important considering financial returns. If this property is not adequately taken into account, risk models are deemed to fail in times of heavy market turmoil when assets become highly correlated, leading to a significant reduction of diversification when it is most needed. 
Finally, our models based on the graphs $\GGG_2,\dots,\GGG_9$ took between 12 hours and 2 days for estimation on a Linux Cluster with $32$ cores. The Gaussian graphical model needs only several minutes for estimation in these dimensions. It is worth noting that other methods for estimating R-vines as the one of \citet{dissmann-etal} failed in these dimensions because of memory consumption or time out.

\section{Discussion}\label{sec:discussion}
We developed a method to estimate ultra high dimensional vine copulas using a novel divide-and-conquer approach inspired by the graphical Lasso. The main idea is to exploit conditional independence for isolated consideration of sub problems of significantly lower dimension compared to the original data. Furthermore, we proposed using graphical independence models also in frameworks of moderate dimension to increase the estimation accuracy of standard vine copula estimation procedures significantly. In all the scenarios, our approach works very well in terms of computation times and penalized goodness of fit measures as GIC, which targets especially sparse datasets. At last, we showed that our approach is of several magnitudes faster than competing methods, however allowing for non-Gaussian dependence. We see further improvement for our contribution with respect to the selection of the corresponding solution of the graphical Lasso, which is the basis for the vine copula model. Currently, we evaluate all these solutions over a grid of penalty values. However, also using criteria as \textit{StARS} \citep{glasso4} seems feasible and worth exploring. Additionally, a more sophisticated value of the currently implemented fill level $k_F = \lceil\log\left(d\right)\rceil$ may be considered, since the current choice is only a rule of thumb motivated by numerical experiments such as presented in Section \ref{sec:numericalexamples}. Finally, a method also worth exploring is to use an \textit{iterated divide and conquer} approach where larger connected components as in $\GGG_{10},\dots,\GGG_{30}$ of the data application can again be divided until all they are all below a given machine-induced threshold. We will consider these aspects as well as the search for applications to non-financial ultra high dimensional datasets in the future.

\section*{Acknowledgement}\label{sec:acknowledgment}
The first author is thankful for a research stipend of the Technische Universit\"at M\"unchen. The second author is supported by the German Research foundation (DFG grant CZ 86/4-1). Numerical computations were performed on a Linux cluster supported by DFG grant INST 95/919-1 FUGG. The \texttt{VineCopula} R-package \citep{vinecopula}, on which our code is based, is greatly acknowledged.
\bibliographystyle{chicago}
\bibliography{bibliography}
\appendix
\newpage
\section{Algorithm \texttt{RVineClusterSelect}}\label{sec:appendix:algorithm}
\IncMargin{0em}
\begin{algorithm}[H]
	\SetAlgoLined
	\SetKwInOut{Input}{input}
	\SetKwInOut{Output}{output}
	\SetKwFunction{Create}{create}
	\SetKwFunction{Assign}{assign}
	\SetKwFunction{Calculate}{calculate}
	\SetKwFunction{Define}{define}
	\SetKwFunction{Order}{order}
	\SetKwFunction{Sample}{sample}
	\SetKwFunction{Return}{return}
	\SetKwFunction{Weight}{weight}
	\SetKwFunction{List}{list}
	\SetKwFunction{Complete}{complete}
	\SetKwFunction{Estimate}{estimate}
	\SetKwFunction{Set}{set}
	\SetKwFunction{Add}{add}
	\SetKwFunction{Delete}{delete}
	\SetKwFunction{Skip}{skip}
	\SetKwFunction{Exit}{exit}
	\SetKwFunction{Select}{select}
	\SetKwFunction{Donothing}{do nothing}
	\SetKwFunction{Compose}{compose}
	\SetKwFunction{Stop}{stop}
	\SetKwFunction{Break}{break}	
	\Input{Data $\bm{X} \in \mathbb{R}^{n \times d}$, $d_T \leq d$, $k_F < d$.}
	\Output{R-Vine in $d$ dimensions.}
	\BlankLine
	\Calculate  $\left(\GGG_1=\left(\bigcup_{i=1}^{p_1}~V_i^1,\bigcup_{i=1}^{p_1}~{E}_i^1\right),\dots,\GGG_J=\left(\bigcup_{i=1}^{p_J}~V_i^J,\bigcup_{i=1}^{p_J}~{E}_i^J\right)\right)$\;
	\Select $\GGG_T$ such that $T = \argmax_{j=1,\dots,J}~\delta_j \leq d_T$ with $\delta_j = \max_{i = 1,\dots,p_i}~\left|V_i^j\right|$\;
	\For{$i = 1$ \KwTo $p_T$}{
		\Set $\nu_i = \left|V_i^T\right|$ \tcp*{dimension of connected component i}
		\For{$\left(j,\ell\right) \in V_i^T$} { 
			\eIf{$\left(j,\ell\right) \notin {E}_i^T$}{
				\Set $c_{j,\ell} = 1$\;}{
				\Estimate pair-copula $c_{j,\ell}$\;}
		}
		\Calculate weights $\mu\left(j,\ell\right) = \mu\left(c_{j,\ell}\right)$ \tcp*{e.\,g.\ Log-Lik., AIC, BIC}
		\Calculate maximum spanning tree $T_1 = \left(V_1 = V_i, E_1\right)$ w.r.t. $\mu$\;
		\For{$k = 2$ \KwTo $\nu_i - 1$ }  {
			\Set $V_k = E_{k-1}$ and $E_{k}$ admissible edges by proximity condition\;
			\For{$\left(j,\ell|\mathbf{D}\right) \in E_k$}{
				\eIf{$\separate{j}{\ell}{\mathbf{D}}{\GGG_T}$}{
					\Set $c_{j,\ell|\mathbf{D}} = 1$\;}{
					\Estimate pair-copula $c_{j,\ell|\mathbf{D}}$\;}
			}
			\Calculate weights $\mu\left(j,\ell|\mathbf{D}\right) = \mu\left(c_{j,\ell|\mathbf{D}}\right)$\;
			\Calculate maximum spanning tree $T_k = \left(V_k, E_k\right)$ w.r.t. $\mu$\;
		}
	}
	\Create R-vine matrix $M$ in $d$ dimensions combining sub-R-vines\;
	\For{$k = 1$ \KwTo $k_F$}{
		\Estimate pair-copulas between sub-R-vines in tree $k$ of in R-vine matrix $M$\;
	}	
	\caption{\texttt{RVineClusterSelect}: Selection of an R-vine in $d$ dimensions.}
	\label{algorithm:clusterselect1}
\end{algorithm}

\newpage
\section{Additional Results for Simulation Study}\label{sec:appendix:sim}
\begin{figure}[H]
	\centering
	\includegraphics[width=0.49\textwidth, trim={0.1cm 0.1cm 0.1cm 0.1cm},clip]{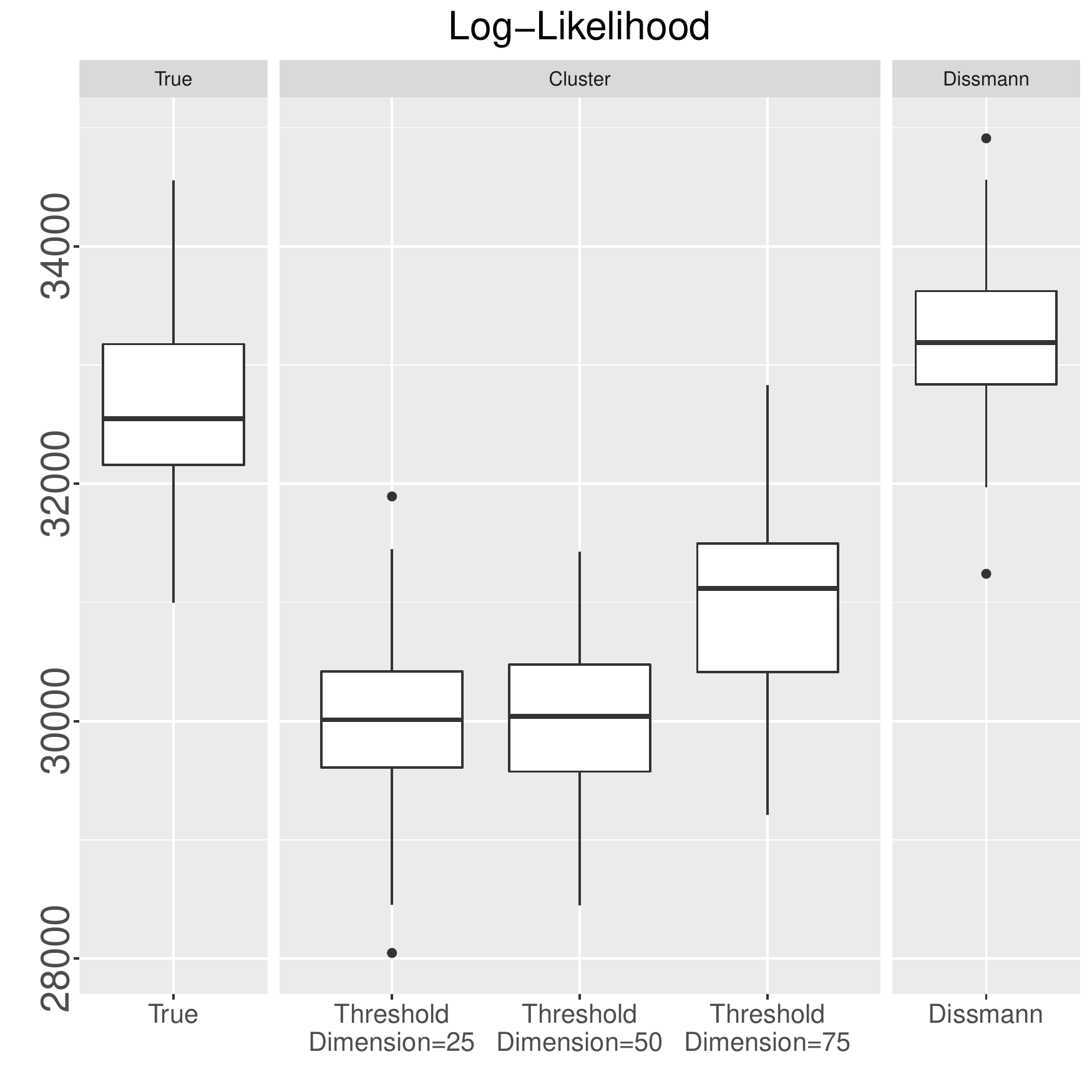}
	\includegraphics[width=0.49\textwidth, trim={0.1cm 0.1cm 0.1cm 0.1cm},clip]{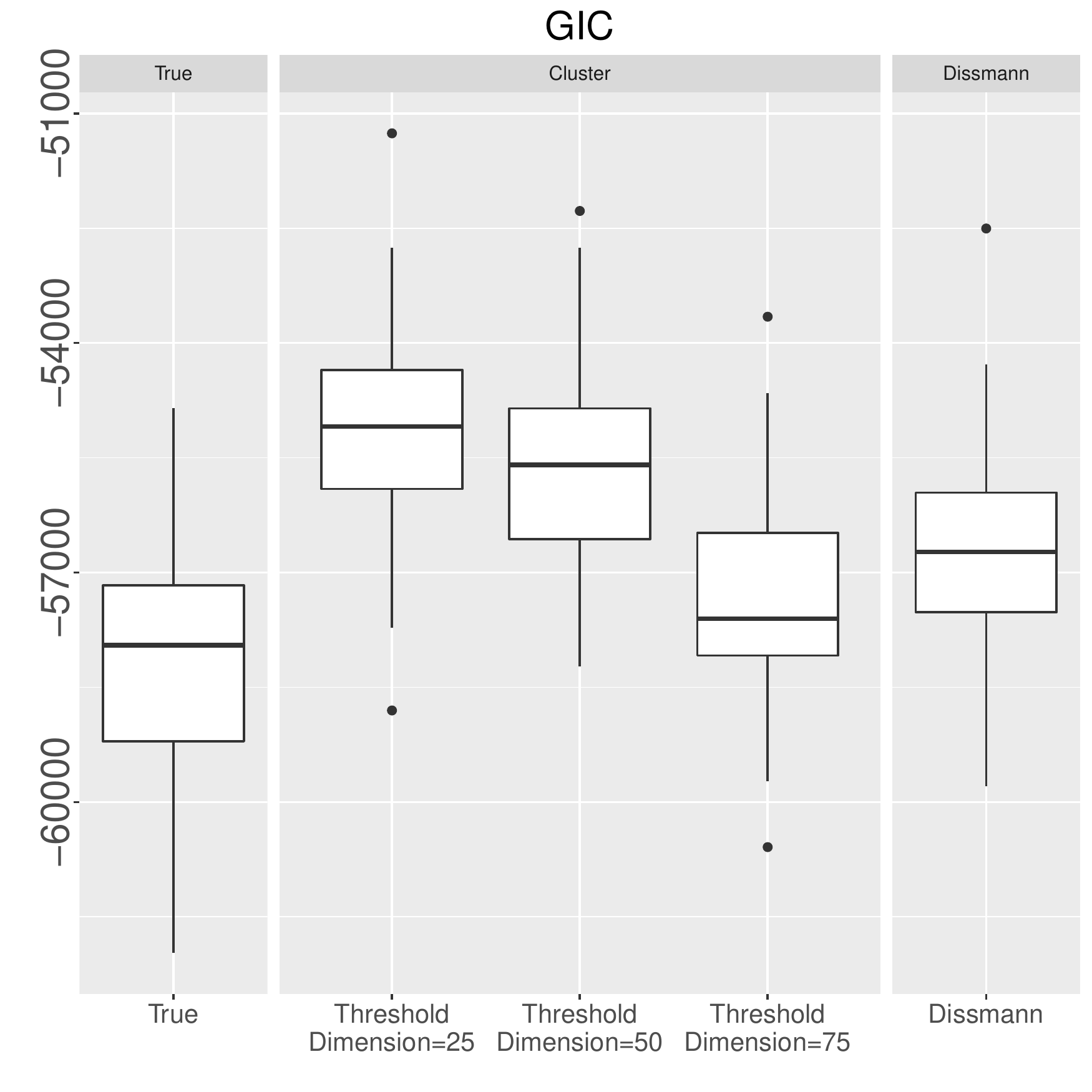}\\
	\includegraphics[width=0.49\textwidth, trim={0.1cm 0.1cm 0.1cm 0.1cm},clip]{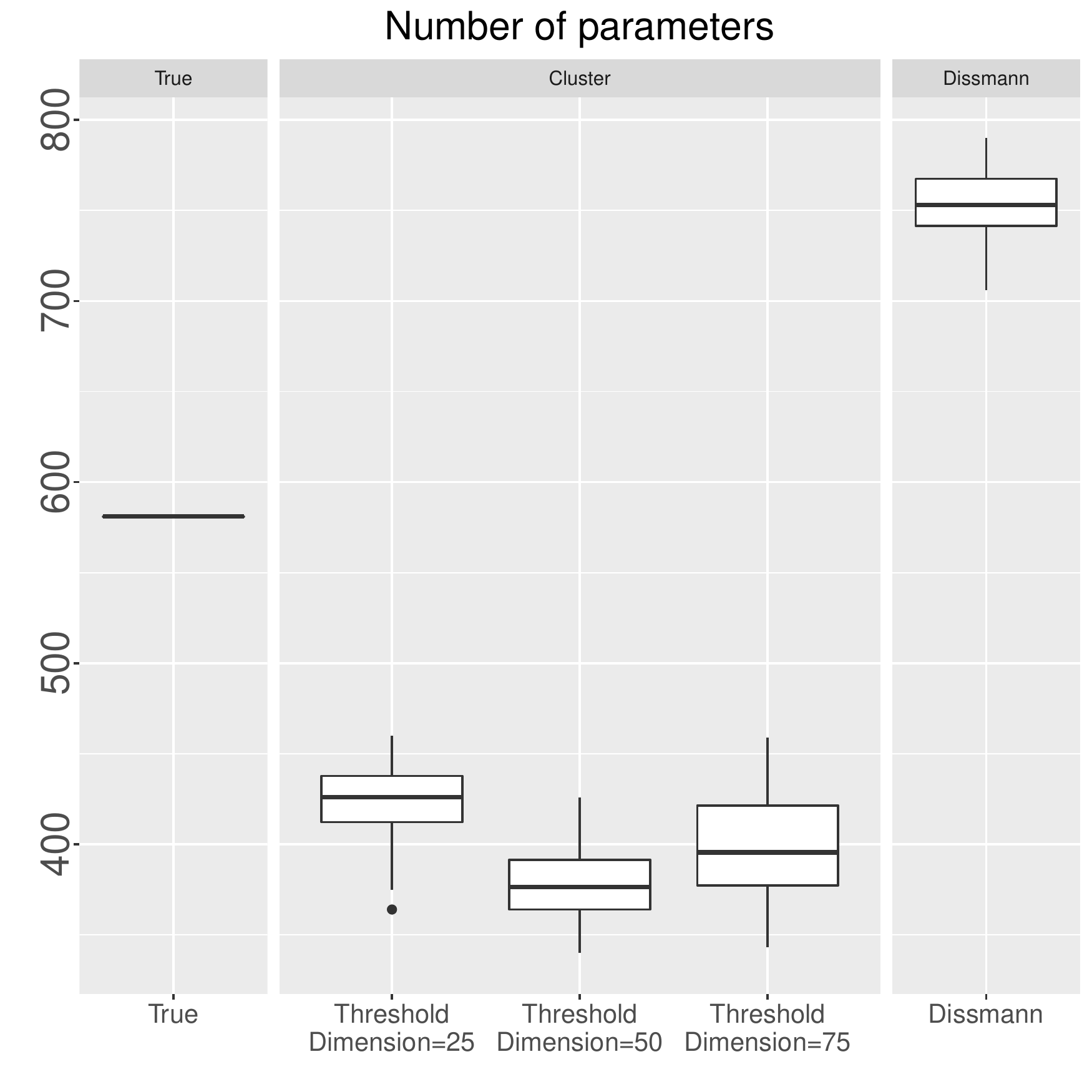}
	\includegraphics[width=0.49\textwidth, trim={0.1cm 0.1cm 0.1cm 0.1cm},clip]{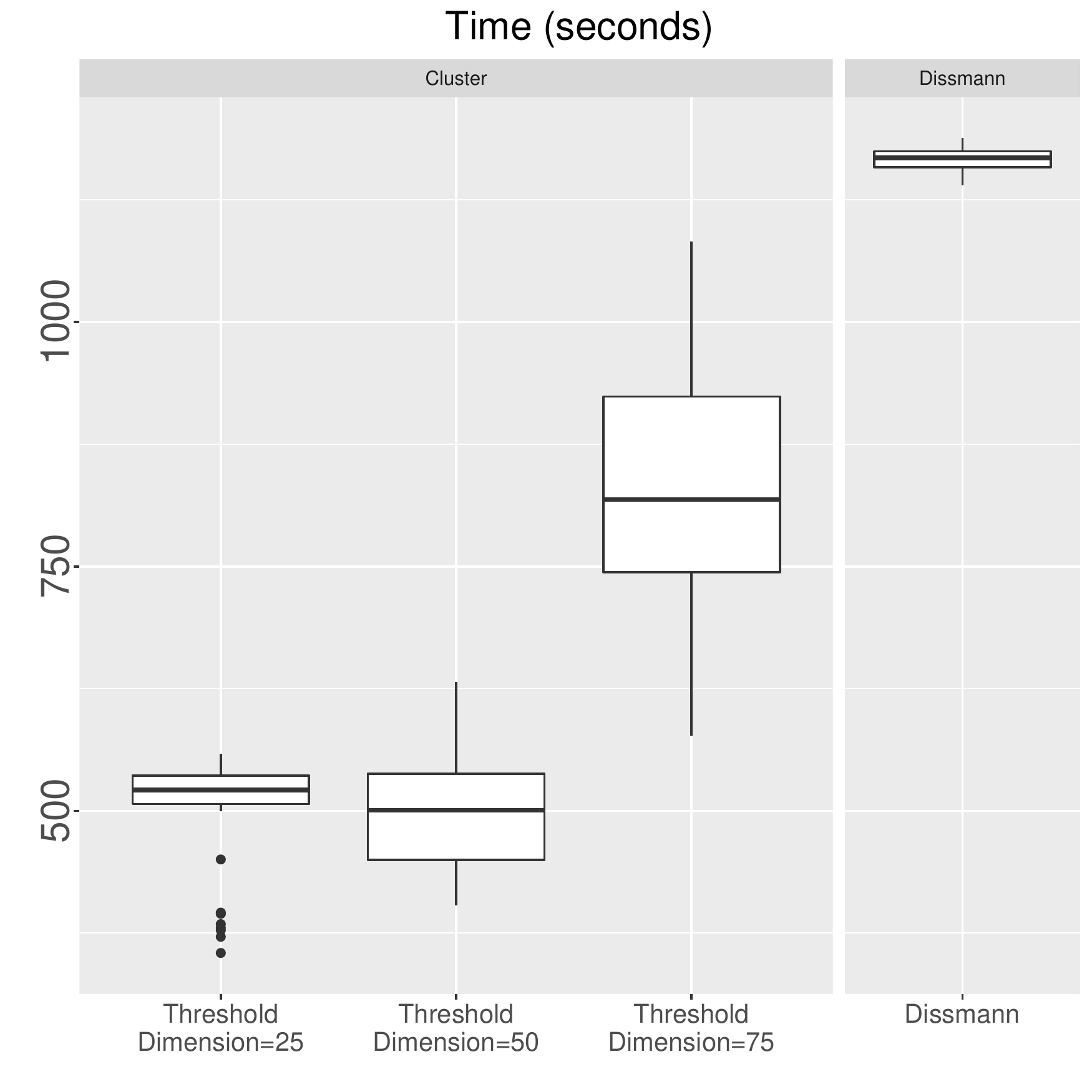}
	\caption{$V_2$: Comparison of \texttt{RVineClusterSelect} algorithm with threshold dimension $d_T = 25,50,75$ and Dissmann's algorithm on \textit{u-scale}: log-likelihood (upper left), GIC (upper right), number of parameters (lower left) and computation time (lower right).}
	\label{fig:simstudy:results5}
\end{figure}

\begin{figure}[H]
	\centering
	\includegraphics[width=0.49\textwidth, trim={0.1cm 0.1cm 0.1cm 0.1cm},clip]{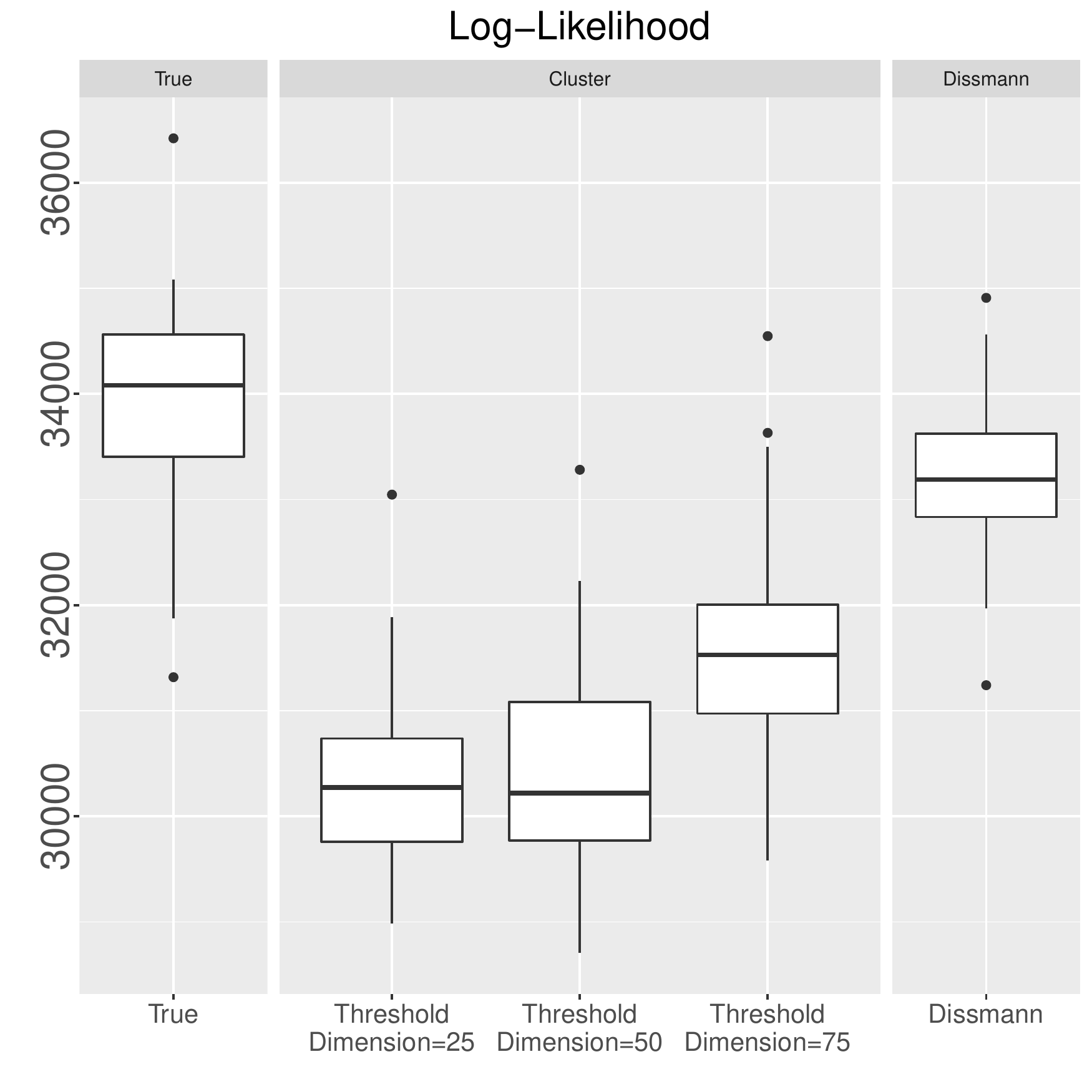}
	\includegraphics[width=0.49\textwidth, trim={0.1cm 0.1cm 0.1cm 0.1cm},clip]{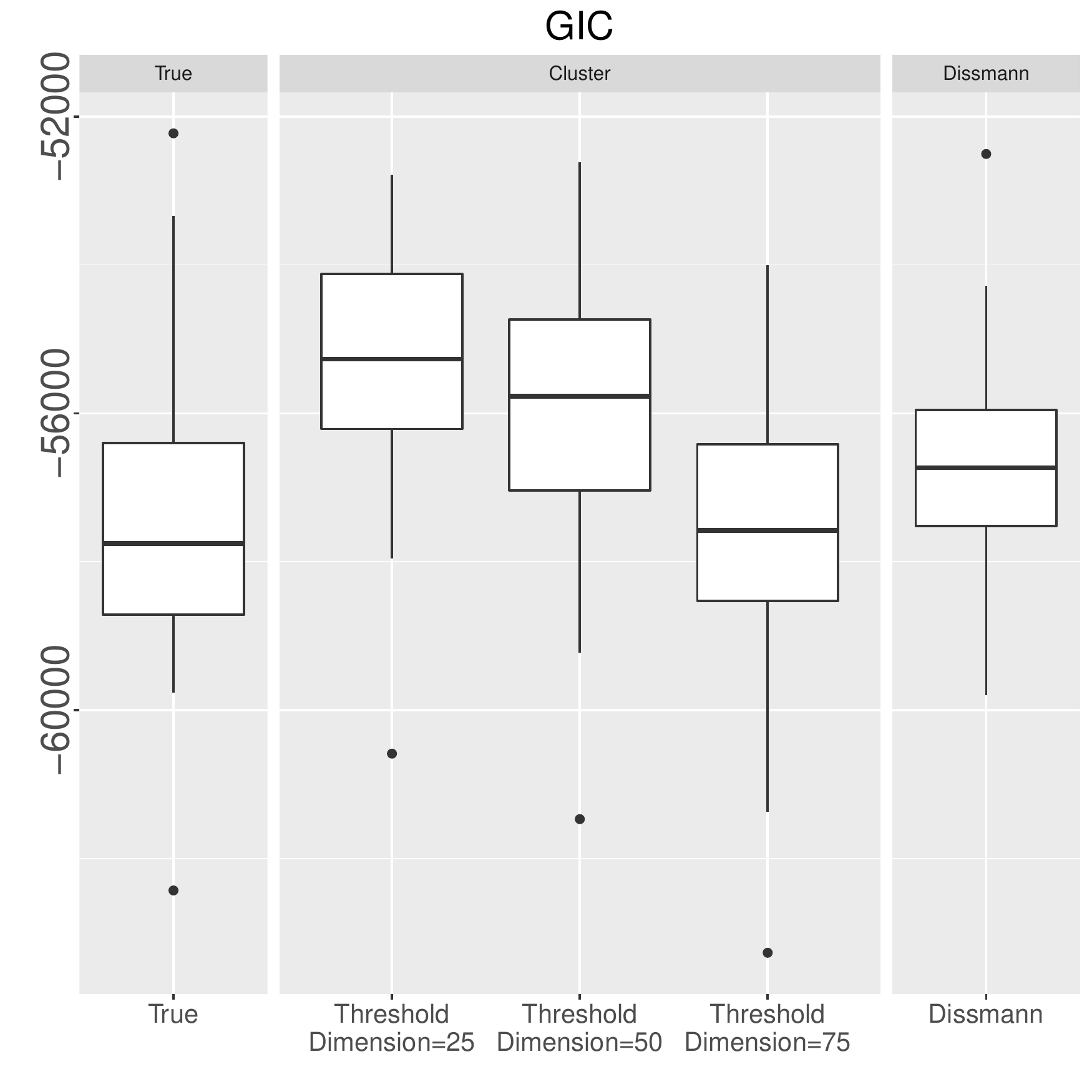}\\
	\includegraphics[width=0.49\textwidth, trim={0.1cm 0.1cm 0.1cm 0.1cm},clip]{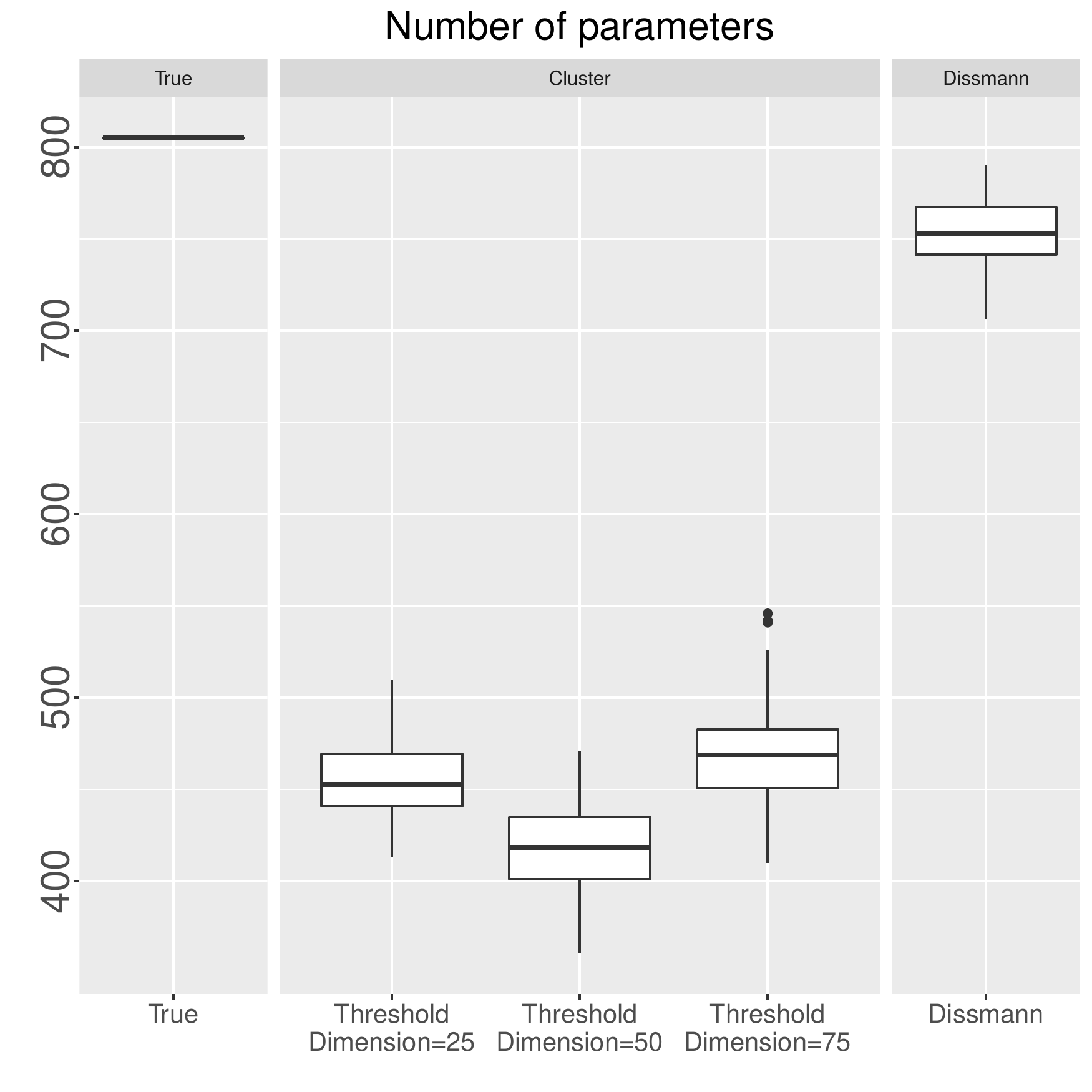}
	\includegraphics[width=0.49\textwidth, trim={0.1cm 0.1cm 0.1cm 0.1cm},clip]{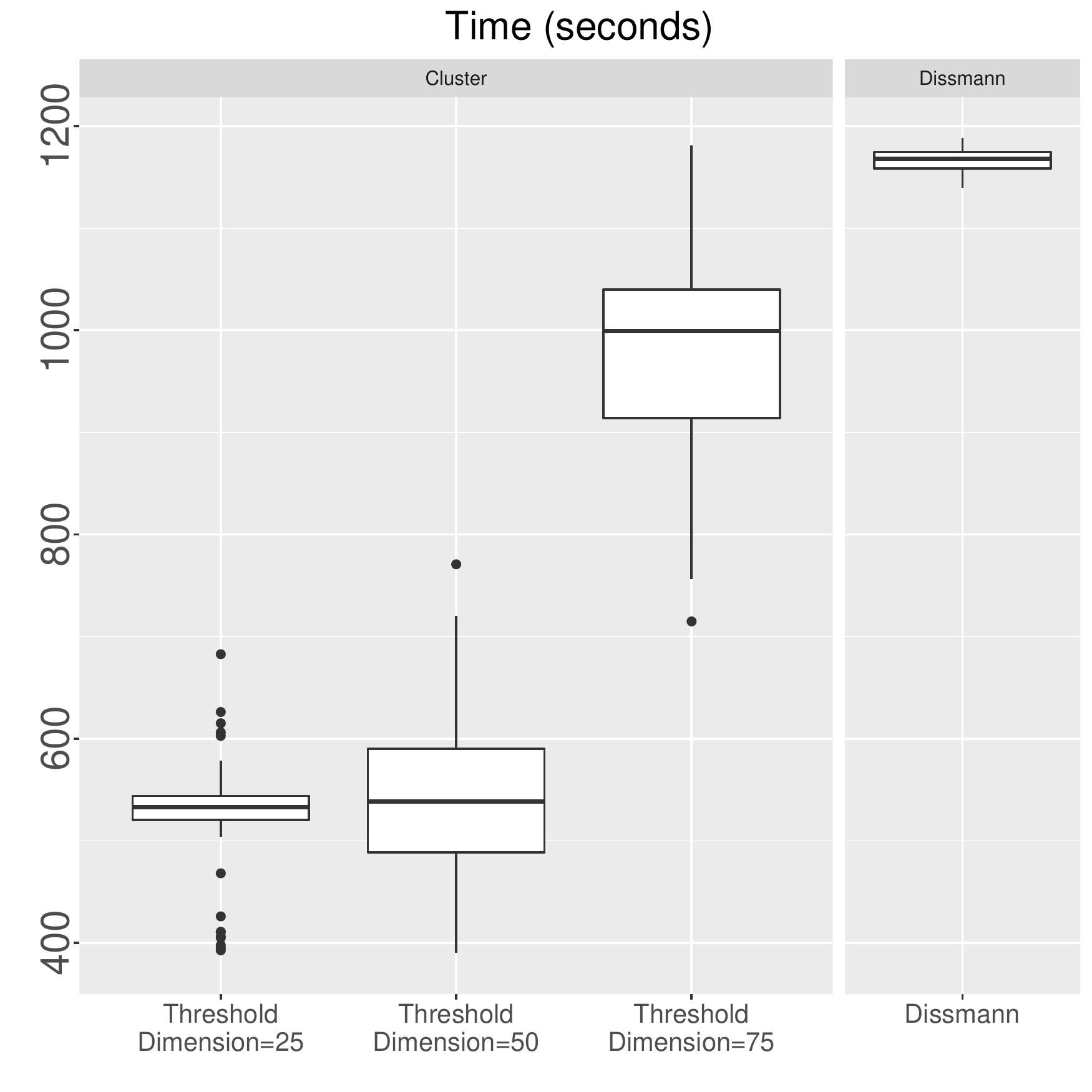}
	\caption{$V_3$: Comparison of \texttt{RVineClusterSelect} algorithm with threshold dimension $d_T = 25,50,75$ and Dissmann's algorithm on \textit{u-scale}: log-likelihood (upper left), GIC (upper right), number of parameters (lower left) and computation time (lower right).}
	\label{fig:simstudy:results10}
\end{figure}

\end{document}